\documentclass[11pt]{article}

\usepackage{epsfig}
\usepackage{amsmath}
\usepackage{amssymb}
\usepackage{fancyhdr}
\usepackage{algorithm}
\usepackage{algorithmic}
\usepackage{graphicx}
\usepackage{hyperref}
\usepackage{cite}
\usepackage{url}

\DeclareMathAlphabet\mathbfcal{OMS}{cmsy}{b}{n}

\begin{document}

\newtheorem{definition}{Definition}
\newtheorem{problem}{Problem}

\begin{center}
\Large \bf
Hybrid Orthogonal Projection and Estimation (HOPE): \\
A New Framework to Probe and Learn Neural Networks
\end{center}

\smallskip

\begin{center}
\large
 \em Shiliang Zhang$^\dagger$ and Hui Jiang$^\ddagger$ \\
\end{center}
\begin{center}
\large
$^\dagger$ National Engineering Laboratory for Speech and Language Information Processing \\
University of Science and Technology of China, Hefei, Anhui, China \\

$^\ddagger$ Department of Electrical Engineering and Computer Science\\
York University,  4700 Keele Street, Toronto, Ontario, M3J 1P3, Canada \\

{\it email: zsl2008@mail.ustc.edu.cn, hj@cse.yorku.ca}
\end{center}


\begin{center}
\bf Abstract
\end{center}

In this paper, we propose a novel model for high-dimensional data, called the {\em Hybrid Orthogonal Projection and Estimation (HOPE)} model, which combines a linear orthogonal projection and a finite mixture model under a unified generative modeling framework. The HOPE model itself can be learned unsupervised from unlabelled data based on the maximum likelihood estimation as well as discriminatively from labelled data. More interestingly, we have shown the proposed HOPE models are closely related to neural networks (NNs) in a sense that each hidden layer can be  reformulated as a HOPE model. As a result, the HOPE framework can be used as a novel tool to probe why and how NNs work, more importantly, to learn NNs in either supervised or unsupervised ways.  In this work, we have investigated the HOPE framework to learn NNs for several standard tasks, including image recognition on MNIST and speech recognition on TIMIT. Experimental results have shown that the HOPE framework yields significant performance gains over the current state-of-the-art methods in various types of NN learning problems, including unsupervised feature learning, supervised or semi-supervised learning. 

\section{Introduction}
\label{sec_intro}

Machine learning systems normally consist of several distinct steps in design, namely feature extraction and data modeling. In feature extraction, some engineering tricks are used to pre-process raw data to extract useful and representative features for the underlying data sets. As a result, this stage is sometimes called {\em feature engineering}. For high-dimensional data, the feature extraction stage needs to distill ``good" features that are representative enough to discriminate different data samples but also it has to perform effective dimensionality reduction to generate less correlated features that can be easily modeled in a lower dimensional space. In data modeling, an appropriate model is selected to model data in the lower-dimensional feature space. There are a wide range of models available for this purpose, such as k-Nearest-Neighbours (kNN) methods, decision trees, linear discriminant models, neural networks, statistical models from the exponential family, or mixtures of the exponential family distributions, and so on. Subsequently, all unknown model parameters are estimated from available training samples based on certain learning criterion, such as maximum likelihood estimation (MLE) or discriminative learning. 

In many traditional machine learning  methods, feature extraction and data modeling are normally conducted independently in two loosely-coupled stages, where feature extraction parameters and model parameters are separately optimized based on rather different criteria. Particularly, feature extraction is normally regarded as a pre-processing stage, where feature extraction parameters are estimated under some quite loose conditions, such as the assumption that data is normally distributed in a high-dimensional space as implied in linear discriminant analysis (LDA) and principal component analysis (PCA). On the other hand, neural networks (NNs) favor an end-to-end learning process, which is normally considered as one exception to the above paradigm. In practice, it has been widely observed that NNs are capable of dealing with almost any type of raw data directly without any explicit feature engineering. In the recent  resurgence of NNs in ``deep learning'', more and more empirical results have demonstrated that deep neural networks (DNNs) can effectively de-correlate high-dimensional raw data and automatically learn useful features from large training sets, without being disturbed by ``the curse of dimensionality''.
However, it still remains as an open question why NNs can handle it and what mechanism is used by NNs to de-correlate high-dimensional raw data to learn good feature representations for many real-world complicated tasks.  

In this paper, we first propose a novel data modeling framework for high-dimensional data, namely Hybrid Orthogonal Projection and Estimation (HOPE). The key argument for the HOPE framework is that feature extraction and data modeling should not be decoupled into two separate stages in learning and a good feature extraction module can not be learned based on some over-simplified and  unrealistic modeling assumptions. The feature extraction and data modeling must be jointly learned and optimized by considering the complex nature of data distributions. This is particularly important in coping with high-dimensional data arising from most real-world applications, such as image, video and speech signals. In the HOPE framework, we propose to model high-dimensional data by combining a relatively simple feature extraction model, namely a linear orthogonal projection, with a powerful statistical model for data modeling, namely a finite mixture model of the exponential family distributions, under a unified generative modeling framework. In this paper, we consider two possible choices for the mixture models, namely Gaussian mixture models (GMMs) and mixtures of the von Mises-Fisher (movMFs) distributions. First of all, an orthogonal linear projection is used in feature extraction to ensure that the highly-correlated high-dimensional raw data is first projected onto a lower-dimensional latent feature space, where all feature dimensions are largely de-correlated. This will give us huge advantages to model data in this feature space rather than the original data space. 
Secondly, in the HOPE framework, we propose to use a powerful model to represent data in the lower-dimensional feature space, rather than using any over-simplified models for computational convenience. This is very important since any real-world data tend to follow a rather complex distribution, which can always be approximated by a finite mixture model up to any arbitrary precision. Thirdly, the most important argument in HOPE is that both the orthogonal projection and the mixture model must be learned jointly according to a single unified criterion. In this paper, we first study how to learn HOPE in an unsupervised manner based on the conventional maximum likelihood (ML) criterion and also explain that the HOPE models can also be learned in a supervised way based on any discriminative learning criterion.

Another important finding in this work is that the proposed HOPE models are closely related to neural networks (NNs) currently  widely used in deep learning. As we will show, any single hidden layer in the most popular rectified linear (ReLU) NNs can always be reformulated as a HOPE model consisting of a linear orthogonal projection and a mixture of von Mises-Fisher 
distributions (movMFs). This formulation helps to explain how NNs actually deal with high-dimensional data and why NNs can de-correlate almost any types of high-dimensional data to generate good feature representations. 
More importantly, this formulation may open up new possibilities to learn NNs more effectively. 
For example, both supervised and unsupervised learning algorithms for the HOPE models can be easily applied to learning NNs. By imposing an explicit orthogonal constraint on the feature extraction layer, we will show that the HOPE methods are very effective in learning NNs for both supervised and unsupervised learning. 
In unsupervised learning, we have shown that the maximum likelihood (ML) based HOPE learning algorithms can serve as a very effective unsupervised learning method to learn NNs from unlabelled data. Our experimental results have shown that the ML-based HOPE learning algorithm can learn good feature representations in an unsupervised way without using any data labels. These unsupervised learned features may be fed to some simple post-stage classifiers, such as linear SVM, to yield comparable performance as  deep NNs supervised learned end-to-end with data labels. Our proposed unsupervised learning algorithms significantly outperform the previous methods based on the Restricted Boltzmann Machine (RBM) \cite{Hinton_2006a} and the autoencoder variants \cite{Bengio07greedylayer-wise,Vincent-2008}.
Moreover, in supervised learning, 
relying on the HOPE models, we have managed to learn some shallow NNs from scratch, which perform comparably with the state-of-the-art deep neural networks (DNNs), as opposed to learn shallow NNs to mimic a pre-trained deep neural network as in \cite{NIPS2014_DNN_no_need_deep}. Finally, the HOPE models can also be used to train deep neural networks and it normally provides significant performance gain over the standard NN learning methods. These results have suggested that the orthogonal constraint in HOPE may serve as a good model regularization in learning of NNs. 

\section{Related Work}
\label{sec_related_work}

Dimensionality reduction in feature space is a well-studied topic in machine learning. Among many, PCA is the most popular technique in this category. PCA is defined as the orthogonal projection of the high-dimensional data onto a lower dimensional linear space, known as the principal subspace, such that the variance of the projected data is maximized \cite{Bishop_PRML_06}. The nice property of PCA is that it can be formulated as an eigenvector problem of the data covariance matrix, where a simple closed-form solution exists. Moreover, in the probabilistic PCA \cite{Tipping99b,Roweis98}, PCA can be expressed as the maximum likelihood solution to a probabilistic latent variable model. In this case, if the projected data are assumed to follow a zero-mean unit-covariance Gaussian distribution in the principal subspace, the probabilistic PCA can also be solved by an exact closed-form solution related to the eigenvectors of the data covariance matrix. The major limitation of PCA is that it is constrained to learn a linear subspace. Many approaches have been proposed to perform nonlinear dimension reduction to learn possible nonlinear manifolds embedded within a high dimensional data space. One way to model the nonlinear structure is through a combination of linear models, so that we make a piece-wise linear approximation to the manifold. This can be obtained by a clustering technique to partition the data set into local groups with standard PCA applied to each group, such as \cite{Hinton_1997,Kambhatla_1997,Tipping_1999}.  
In \cite{Tipping_1999}, the high-dimensional raw data is assumed to follow a mixture model, where each component may perform its own maximum likelihood PCA in a local region. However, in these methods, it may be quite challenging to perform effective clustering or estimate good mixture models for high-dimensional data due to the strong correlation in various data dimensions. Alternatively, in \cite{Hinton_2006}, a flexible nonlinear method is proposed to reduce feature dimension based on a deep auto-associative neural network. 

Similar to PCA, the Fisher's linear discriminant analysis (LDA) can also be viewed as a linear dimensionality reduction technique. However, PCA is unsupervised in the sense that PCA depends only on the data while Fisher's LDA is supervised since it uses both data and  class-label information. The high-dimensional data are linearly projected to a subspace where various classes are best distinguished as measured by the Fisher criterion. In \cite{Kumar98}, the so-called heteroscedastic discriminant analysis (HDA) is proposed to extend LDA to deal with high-dimensional data with heteroscedastic covariance, where a linear projection can be learned from data and class labels based on the maximum likelihood criterion. 


\section{Hybrid Orthogonal Projection and Estimation (HOPE)}
\label{sec_hope}

Consider a standard PCA setting, where each data sample is represented as a high-dimensional vector ${\bf x}$ with dimensionality $D$. Our goal is to learn a linear projection, represented as a matrix ${\bf U}$, to map each data sample onto a space having dimensionality $M<D$, which is called the latent feature space hereafter in this paper.
Our proposed HOPE model is essentially a generative model in nature but it may also be viewed as a generalization to extend the probabilistic PCA in \cite{Tipping99b} to consider a complex data distribution that has to be modeled by a finite mixture model in the latent feature space. This setting is very different from \cite{Tipping_1999}, where 
the original data ${\bf x}$ is modeled by mixture models in the original higher $D$-dimensional raw data space.

\subsection{HOPE: combining generalized PCA with generative model}

Assume we have a full-size  $D \times D$ orthogonal matrix $\hat{\bf U}$, satisfying  $\hat{\bf U}^{T} \hat{\bf U}   = \hat{\bf U} \hat{\bf U}^{T} = {\bf I}$, each data sample ${\bf x}$ in the original $D$-dimensional data space  can be decomposed based on all orthogonal row vectors of $\hat{\bf U}$, denoted as ${\bf u}_i$ with $i=1, \cdots, D$, as follows:

\begin{equation}
{\bf x}  =  \sum_{i=1} ^D  \;\; ({\bf x} \cdot {\bf u}_i) \, {\bf u}_i.
\end{equation}

As shown in PCA, each high-dimensional data ${\bf x}$ can normally be represented fairly precisely in a lower-dimensional principal subspace and the contributions from the remaining dimensions may be viewed as random residual noises that have sufficiently small variances. 
Therefore, we have
\begin{equation} \label{eq-HOPE-signals}
{\bf x}  = \underbrace{ ({\bf x} \cdot {\bf u}_1) \, {\bf u}_1 + \cdots + ({\bf x} \cdot {\bf u}_M) \, {\bf u}_M}_
{signal \;\; component \;\; \tilde{\bf x}}
+ \underbrace{ ({\bf x} \cdot {\bf u}_{M+1}) \, {\bf u}_{M+1} + \cdots  + ({\bf x} \cdot {\bf u}_{D}) \, 
{\bf u}_{D}  }_{noise \;\; component\;\; \bar{\bf x}}
\end{equation}

Here we are interested in learning an $M \times D$ projection matrix, denoted as ${\bf U}$, to extract the signal component $\tilde{\bf x}$. First of all, if $M$ $(M<D)$ is selected properly, the projection may serve as an effective feature extraction for signals as well as a mechanism to  eliminate unwanted noises from the higher dimensional raw data. This may make the subsequent learning process more robust and less prone to overfitting. Secondly, all $M$ row vectors ${\bf u}_i$ with $i=1, \cdots, M$ are learned to represent signals well in a lower $M$-dimension space. Furthermore, since all ${\bf u}_i$ are orthogonal, it implies the latent features are largely de-correlated.
This will significantly simplify the following learning problem as well. 

In this case, each $D$-dimension data  sample, ${\bf x}$, is linearly projected onto an $M$-dimension vector ${\bf z}$ as 
${\bf z} =  {\bf U}  {\bf x}$,  where ${\bf U}$ is an orthogonal matrix, satisfying $ {\bf U} {\bf U}^{T}= {\bf I}$. Meanwhile, 
we denote the projection of the unwanted noise component $\bar{\bf x}$ as ${\bf n}$, and  
${\bf n}$ can be similarly computed as ${\bf n} =  {\bf V}  {\bf x}$, where ${\bf V}$ is another orthogonal matrix corresponding to all noise dimensions, satisfying $ {\bf V} {\bf V}^{T}= {\bf I}$. Moreover, ${\bf V}$ is orthogonal to  ${\bf U}$, i.e.
${\bf V} {\bf U}^T = {\bf 0}$. 
In overall, we may represent the above projection as follows:

\begin{equation}
\big[ {\bf z}  ; \;{\bf n} \big] =   \big[ {\bf U} ; \; {\bf V}  \big]  {\bf x} = \hat{\bf U}  {\bf x}
\end{equation}
where $\hat{\bf U}$ is the above-mentioned $D\times D$ orthogonal projection matrix. 

Moreover, it is straightforward to show that the signal component $\tilde{\bf x}$ and the residual noise $\bar{\bf x}$ in the original data space can be easily computed from the above projections as follows:

\begin{equation}
\tilde{\bf x} = {\bf U}^T {\bf z}  =  {\bf U}^T {\bf U}  {\bf x}  
\end{equation}

\begin{equation}
\bar{\bf x}  = {\bf x} - \tilde{\bf x} =  ({\bf I} - {\bf U}^T {\bf U} ) {\bf x}
\end{equation}

In the following, we consider how to learn the projection matrix ${\bf U}$ to 
represent $D$-dimensional data well in a lower $M$-dimension feature space.
If this projection is learned properly, we may assume 
the above signal projection, ${\bf z}$, and the residual noise projection, ${\bf n}$, are independent in the latent feature space. Therefore, we may derive the probability distribution of the original data as follows:
\begin{equation} \label{eq-pdf-two-spaces}
p({\bf x}) = |\hat{\bf U}^{-1}| \cdot p({\bf z}) \cdot p({\bf n})
\end{equation}
where $\hat{\bf U}^{-1}$ denotes the Jacobian matrix to linearly map data from the projected space back to the original data space. If $\hat{\bf U}$ is orthonormal, the above Jacobian term equals to one. 
In this work, 
we follow \cite{Tipping99b} to assume the residual noise projection ${\bf n}$ follows an isotropic covariance Gaussian distribution in the (D-M)-dimensional space, i.e. $p({\bf n})  \; \sim \; {\cal N}({\bf n} \; | \; {\bf 0}, \sigma^2 {\bf I})$,
\footnote{Without losing generality, we may simply normalize the training data to ensure that the residual noises have zero mean.} where $\sigma^2$ is a variance parameter to be learned from data. As for the signal projection ${\bf z}$, we adopt a quite different approach, as described below in detail.

In all previous works, the signal projection ${\bf z}$ is assumed to follow a simple distribution in the $M$-dimension space. For example, ${\bf z}$ is assumed to follow a zero-mean unit-covariance Gaussian distribution in probabilistic PCA \cite{Tipping99b,Roweis98}. 
The advantage of this assumption is that an exact closed-form solution may be derived to calculate the projection matrix ${\bf U}$ using the spectral methods based on the data covariance matrix.  

However, in most real-world applications, ${\bf z}$ still locates in a very high-dimensional space even after the linear projection, it does not make sense to assume ${\bf z}$ follows a simple unimodal distribution. As widely reported in the literature, it is empirically observed that real-world  data normally do not follow a unimodal distribution in a high-dimensional space and they usually appear only in multiple concentrated regions in the high-dimensional space. More realistically, it is better to assume that ${\bf z}$ follows a finite mixture model in the $M$-dimension feature space
because a finite mixture model may theoretically approximate any arbitrary statistical distribution as long as a sufficiently large number of mixture components are used. For simplicity, we may assume ${\bf z}$ follows a finite mixture of some exponential family distributions:
\begin{equation} \label{eq-HOPE-mixture}
p({\bf z}) 
= \sum_{k=1} ^K \;\; \pi_k \cdot f_k({\bf z} | {\boldsymbol \theta}_k)
\end{equation}
where $\pi_k$ denotes mixture weights with $\sum_{k=1}^K \pi_k = 1$, and $f_k({\bf z} | {\boldsymbol \theta}_k)$ stands for a unimodal distribution from the exponential family with model parameters ${\boldsymbol \theta}_k$. We use ${\boldsymbol \Theta}$ to denote all model parameters in the mixture model, i.e., ${\boldsymbol \Theta} = \{ {\boldsymbol \theta}_k, \pi_k \; | \; k = 1, \cdots, K\} $.
In practice, $f_k({\bf z} | {\boldsymbol \theta}_k)$ is chosen from the exponential family based on the nature of data. In this paper, we consider two possible choices for high-dimensional continuous data, namely the multivariate Gaussian distributions and the von Mises-Fisher (vMF) distributions. The learning algorithm can be easily extended to other models in the exponential family. 

For example, if we choose the multivariate Gaussian distribution, then ${\bf z}$ follows a Gaussian mixture model  (GMM) as follows:
\begin{eqnarray}
p({\bf z}) 
= \sum_{k=1} ^K \;\; \pi_k \cdot f_k({\bf z} | {\boldsymbol \theta}_k)
= \sum_{k=1} ^K \;\; \pi_k \cdot {\cal N}( {\bf z} \; | \;{\boldsymbol \mu}_k, \Sigma_k)
\end{eqnarray}
where ${\cal N}( {\bf z} \; | \;{\boldsymbol \mu}_k, \Sigma_k)$ denotes a multivariate Gaussian distribution with the mean vector ${\boldsymbol \mu}_k $ and the covariance matrix $\Sigma_k$.  Since the projection matrix ${\bf U}$ is orthogonal, all dimensions in ${\bf z}$ are highly de-correlated. Therefore, it is reasonable to assume each Gaussian component has a diagonal covariance matrix $\Sigma_k$. This may significantly simplify the model estimation of GMMs. 

Alternatively, we may select a less popular model, i.e., the von Mises-Fisher (vMF) distribution.\footnote{The main reason why we are interested in the von Mises-Fisher (vMF) distributions is that the choice of the vMF model can strictly link our HOPE model to regular neural networks in deep learning. We will elucidate this later in this paper. } The vMF distribution may be viewed as a generalized normal distribution defined  on a high-dimensional spherical surface. In this case, ${\bf z}$ follows a mixture of the von Mises-Fisher (movMF) distributions as follows:
\begin{eqnarray}  \label{eq-movMF-pdf}
p({\bf z}) = 
\sum_{k=1} ^K \;\; \pi_k \cdot f_k({\bf z} | {\boldsymbol \theta}_k) 
= \sum_{k=1} ^K \;\; \pi_k \cdot {\cal C}_{M} (|{\boldsymbol \mu}_k|) \cdot e^{{\bf z} \cdot {\boldsymbol \mu}_k}    
\end{eqnarray}
where ${\bf z}$ is located on the surface of an M-dimensional sphere, i.e., $|{\bf z}| =1$, 
${\boldsymbol \mu}_k$ denotes all model parameters of the $k$-th vMF component and it is an $M$-dimensional vector in this case, and
${\cal C}_{M} (\kappa)$ is the probability normalization term of the $k$-th vMF component defined as:
\begin{equation}
{\cal C}_{M} (\kappa) = \frac{\kappa^{M/2-1} }{(2 \pi)^{M/2} I_{M/2-1}(\kappa)}
\end{equation}
where $I_v(\cdot)$ denotes the modified Bessel function of the first kind at order $v$.

\section{Unsupervised Learning of HOPE Models}
\label{sec-upsupervised-MLE-HOPE}

Obviously, the HOPE model is essentially a generative model that combines feature extraction and data modelling together, and thus its model parameters, including both the projection matrix and the mixture model,  can be estimated based on the maximum likelihood (ML) criterion, just like normal generative models as well as the probabilistic PCA in \cite{Tipping99b}.
However, since ${\bf z}$ follows a mixture distribution,  no closed-form solution is available to derive either the projection matrix or the mixture model. In this case, some iterative optimization algorithms, such as stochastic gradient descent (SGD) \cite{Bottou04, Bottou11}, may be used to jointly estimate both the projection matrix ${\bf U}$ and the underlying mixture model altogether to maximize a joint likelihood function. In this section, we assume the projection matrices, not only ${\bf U}$ but also  the whole $\hat{\bf U}$, are all orthonormal.  As a result, the Jacobian term in eq.(\ref{eq-pdf-two-spaces}) disappears since it equals to one. Refer to Appendix \ref{appendix-non-orthonormal-U} for the case where ${\bf U}$ is not orthonormal.

Given a training set is available as $ {\bf X} = \{{\bf x}_n \; | \; n=1, \cdots, N \}$, 
assume that all ${\bf x}_n$ are normalized to be of unit length as $|{\bf x}_n|=1$,
the joint log-likelihood function related to all HOPE parameters, including the projection matrix ${\bf U}$, the mixture model ${\boldsymbol \Theta} = \{ {\boldsymbol \theta}_k | k=1,\cdots, K \}$ and residual noise variance $\sigma$, can be expressed 
as  follows:

\begin{eqnarray}  \label{eq-HOPE-likelihood-WithoutJacobian}
{\cal L}({\bf U}, {\boldsymbol \Theta}, \sigma \; | \; {\bf X}) & = & 
 \sum_{n=1}^N \; \ln \Pr({\bf x}_n) = \sum_{n=1}^N  \bigg[ \ln \Pr({\bf z}_n) + \ln \Pr({\bf n}_n) \bigg] \nonumber \\ 
& = & \underbrace{ \sum_{n=1}^N \; \ln \left(\sum_{k=1}^K  \;\; \pi_k \cdot f_k( {\bf U}  {\bf x}_n | {\boldsymbol \theta}_k) \right)}_{{\cal L}_1({\bf U},{\boldsymbol \Theta})}
+ \underbrace{\sum_{n=1}^N \;  \ln \bigg( {\cal N}\big( {\bf n}_n \, | \, {\bf 0}, \sigma^2 {\bf I}     \big)   \bigg)}_{{\cal L}_2({\bf U}, \sigma)} \nonumber \\
\end{eqnarray}

The HOPE parameters, including ${\bf U}$, ${\boldsymbol \Theta}$ and $\sigma$, can all be estimated by maximizing the above likelihood function as:
\begin{equation}
\{ {\bf U}^*, {\boldsymbol \Theta}^*, \sigma^*\}	  = {\arg\max}_{ {\bf U}, {\boldsymbol \Theta}, \sigma } \;\;\; 
{\cal L}({\bf U}, {\boldsymbol \Theta}, \sigma \; | \; {\bf X})
\end{equation}
subject to the orthogonal constraint: 
\begin{equation} \label{eq-HOPE-orthogonal-constraint-isolated}
{\bf U} {\bf U}^{T}  = {\bf I}.
\end{equation}

There are many methods to enforce the above orthogonal constraint in the optimization. For example, we may periodically apply the Gram-Schmidt process in linear algebra to orthogonalize ${\bf U}$ during the optimization process. In this work, for computational efficiency, we follow \cite{Bao2013} to cast the orthogonal constraint condition in eq.(\ref{eq-HOPE-orthogonal-constraint-isolated}) as a penalty term in the objective function to convert the above constrained optimization problem into an unconstrained one as follows:
\begin{equation}\label{HOPE-unconstrained-obj}
	\{ {\bf U}^*, {\boldsymbol \Theta}^*, \sigma^* \}	  = {\arg\max}_{ {\bf U}, {\boldsymbol \Theta}, \sigma } \;\;\;
	\bigg[ {\cal L}({\bf U}, {\boldsymbol \Theta}, \sigma \; | \; {\bf X}) - \beta \cdot {\cal D}({\bf U}) \bigg]
\end{equation}
where $\beta$ ($\beta>0$) is a control parameter to balance the contribution of the penalty term, and the penalty term $ {\cal D}({\bf U})$ is a differentiable function as:
\begin{equation} \label{eq-HOPE-orthogonal-constraint-term}
{\cal D}({\bf U}) = \sum_{i=1}^{M} \sum_{j=i+1}^{M} \frac{|{\bf u}_{i}\cdot {\bf u}_{j}|}{|{\bf u}_{i}| \cdot |{\bf u}_{j}|}
\end{equation}
with ${\bf u}_{i}$ denoting the $i$-th row vector of the projection matrix ${\bf U}$, and 
${\bf u}_{i}\cdot {\bf u}_{j}$ representing the inner product of ${\bf u}_{i}$ and ${\bf u}_{j}$.
The norms of all row vectors of ${\bf U}$ need to be normalized to one after each update. 

In this work, we propose to use the stochastic gradient descent (SGD) method to optimize the objective function in eq.(\ref{HOPE-unconstrained-obj}). In this case, given any training data or a mini-batch of them, we calculate the gradients of the objective function with respect to the projection matrix, ${\bf U}$, and the parameters of the mixture model, ${\boldsymbol \Theta}$, and then update them iteratively until the objective function converges. The gradients of ${\cal L}_1({\bf U}, {\boldsymbol \Theta})$ depends on the mixture model to be used. In the following, we first consider how to compute the derivatives of ${\cal D}({\bf U})$ and ${\cal L}_2({\bf U}, \sigma)$, which are general for all HOPE models. After that, as two examples, we will show how to calculate the derivatives of ${\cal L}_1({\bf U}, {\boldsymbol \Theta})$ for GMMs and movMFs. 

\subsection{Dealing with the penalty term ${\cal D}({\bf U})$}

Following \cite{Bao2013}, the gradients of the penalty term ${\cal D}({\bf U})$ with respect to each row vector, ${\bf u}_{i}$ ($i=1, \cdots, M$), can be easily derived as follows:
\begin{equation} \label{eq-HOPE-orthogonal-constraint-derivatives}
\frac{\partial {\cal D}({\bf U}) }{ \partial {\bf u}_i } 
= \sum\limits_{j = 1}^M \;\; { g_{ij} } \cdot \left[ \frac{ {\bf u}_j } { {\bf u}_i  \cdot {\bf u}_j } 
- \frac{ {\bf u}_i } {  {\bf u}_i \cdot {\bf u}_i} \right]
\end{equation}
where $g_{ij}$ denotes the absolute cosine value of the angle between two row vectors, ${\bf u}_i$ and  ${\bf u}_j$, computed as follows:
\begin{equation} \label{eq-G-matrix-elements}
 g_{ij} = \frac{|{\bf u}_{i} \cdot {\bf u}_{j}|}{| {\bf u}_{i}|  \cdot |{\bf u}_{j}|}.
\end{equation}

The above derivatives can be equivalently represented as the following matrix form:
\begin{equation}   \label{eq-derivatives-maxtrix-form}
\frac{\partial {\cal D}({\bf U}) }{ \partial {\bf U} } = ( {\bf D} - {\bf B}) {\bf U} 	
\end{equation}
where  ${\bf D}$ is an  $M \times M$ matrix, with its elements computed as 
 $d_{ij} = \frac{\mbox{sign}({\bf u}_i  \cdot {\bf u}_j)}{{|\bf u}_i|  \cdot |{\bf u}_j| } \; (1\leq i,j \leq M)$,
and ${\bf B}$ is an $M \times M$ diagonal matrix, with its diagonal elements computed as 
$b_{ii} = \frac{\sum_{j=1} g_{ij}}{{\bf u}_i  \cdot {\bf u}_i} \; (1 \leq i \leq M)$.
 
\subsection{Dealing with the noise model term ${\cal L}_2$}

The log-likelihood function related to the noise model, ${\cal L}_2({\bf U}, \sigma)$,
can be expressed as:
\begin{equation}
{\cal L}_2({\bf U}, \sigma) = - \frac{N(D-M)}{2} \ln (\sigma^2) - \frac{1}{2 \sigma^2} \sum_{n=1}^N \;
 {\bf n}_n^T {\bf n}_n.
\end{equation}

And we have:
\begin{eqnarray}
{\bf n}_n^T {\bf n}_n & = & ({\bf x}_n - {\bf U}^T {\bf z}_n)^T ({\bf x}_n - {\bf U}^T {\bf z}_n)
= ({\bf x}_n - {\bf U}^T {\bf U} {\bf x}_n)^T ({\bf x}_n - {\bf U}^T {\bf U} {\bf x}_n) \nonumber \\
& = & {\bf x}_n^T {\bf x}_n - 2 {\bf x}_n^T  {\bf U}^T {\bf U} {\bf x}_n
+  {\bf x}_n^T  {\bf U}^T {\bf U} {\bf U}^T {\bf U}    {\bf x}_n
\end{eqnarray}

Therefore, ${\cal L}_2({\bf U}, \sigma)$ can be expressed as:
\begin{equation}
{\cal L}_2({\bf U}, \sigma) = - \frac{N(D-M)}{2} \ln (\sigma^2) - \frac{1}{2 \sigma^2} \sum_{n=1}^N \;
\bigg[ {\bf x}_n^T {\bf x}_n - 2 {\bf x}_n^T  {\bf U}^T {\bf U} {\bf x}_n
+  {\bf x}_n^T  {\bf U}^T {\bf U} {\bf U}^T {\bf U}    {\bf x}_n \bigg]
\end{equation}

The gradient with respect to ${\bf U}$ \footnote{We may use the constraint ${\bf U} {\bf U}^T= {\bf I}$ to significantly simplify the above derivation. However, that leads to a gradient computation strongly relying on the orthogonal constraint. Since we use SGD to iteratively optimize all model parameters, including  ${\bf U}$. We can not ensure ${\bf U} {\bf U}^T= {\bf I}$ strictly holds anytime in the SGD process. Therefore, the simplified gradient usually yields poor convergence performance. }
can be derived as follows:

\begin{eqnarray}
\frac{\partial {\cal L}_2({\bf U}, \sigma)}{\partial {\bf U}}	& = &   \frac{1}{\sigma^2}  \sum_{n=1}^N \; 
\bigg[2 {\bf U} {\bf x}_n {\bf x}^T_n  -  {\bf U} {\bf U}^T {\bf U} {\bf x}_n {\bf x}^T_n   - 
{\bf U} {\bf x}_n {\bf x}_n^T {\bf U}^T {\bf U} \bigg]
\nonumber \\
& = &  \frac{1}{\sigma^2}  \sum_{n=1}^N \; {\bf U}  \bigg[ {\bf x}_n (\bar{\bf x}_n)^T
+ \bar{\bf x}_n ({\bf x}_n)^T  \bigg].
\end{eqnarray}

%
%
%
%

For the noise variance $\sigma^2$, we can easily derive the following closed-form update formula by vanishing its derivative to zero:

\begin{equation} \label{eq-noise-variance-update}
\sigma^2 =  \frac{1}{N(D-M)}\sum_{n=1}^N \; {\bf n}_n^T {\bf n}_n
\end{equation}

As long as the learned noise variance $\sigma^2$ is small enough, maximizing the above term ${\cal L}_2$ will force all signal dimensions into the projection matrix ${\bf U}$ and only the residual noises will be modelled by ${\cal L}_2$.

\subsection{Computing ${\cal L}_1$ for GMMs}

In this section, we consider how to compute the partial derivatives of  ${\cal L}_1({\bf U}, {\boldsymbol \Theta})$ for  GMMs. Assume each mini-batch ${\bf X}$ consists of a small subset of randomly selected training samples, ${\bf X} = \{ {\bf x}_n \; | \; n=1, \cdots, N \}$, the log likelihood function of HOPE models with GMMs can be represented as follows:
\begin{equation} \label{GMM_HOPE}
{\cal L}_1({\bf U}, {\boldsymbol \Theta} ) =
 \sum\limits_{n = 1}^N \;\; \ln \left[ \sum\limits_{k = 1}^K \;
 {\pi _k} \cdot {\cal N}( {\bf U} {\bf x}_n \;|\;{{\boldsymbol \mu}_k},{\Sigma _k})   \right]
\end{equation}

The partial derivative of ${\cal L}_1 ({\bf U}, {\boldsymbol \Theta})$  w.r.t the mean vector, ${\boldsymbol \mu}_k$, of the $k$-th Gaussian component can be calculated as follows:
\begin{equation}
\frac{{\partial {\cal L}_1({\bf U}, {\boldsymbol \Theta} )}}{{\partial {\boldsymbol \mu _k}}} =
\sum\limits_{n = 1}^N {\gamma_k ({\bf z}_n) \cdot \Sigma _k^{ - 1} ({ \mathbf{z}_n} - {\boldsymbol \mu _k})}  
\end{equation} 
where ${\bf z}_n = {\bf U} {\bf x}_n$, and $\gamma_k({\bf z}_n)$ denotes the so-called occupancy statistics of the $k$-th Gaussian component, computed as 
$\gamma_k({\bf z}_{n}) =  {\frac{{{\pi _k} {\cal N}({\mathbf{z}_n}|{\boldsymbol \mu _k},{\Sigma _k})}}{{\sum\limits_{j = 1}^K {{\pi _j} {\cal N}({\mathbf{z}_n}|{\boldsymbol \mu _j},{\Sigma _j})} }}}$.

The partial derivative of ${\cal L}_1({\bf U}, {\boldsymbol \Theta} )$ w.r.t  $\pi_k$ can be simply derived as follows:
\begin{equation}
   \frac{\partial {\cal L}_1({\bf U}, {\boldsymbol \Theta} ) }{\partial \pi_k} = \sum_{n=1}^{N}  \;\; \frac{\gamma_k({\bf z}_{n}) }{ \pi_k}
\end{equation}

The partial derivative of ${\cal L}_1({\bf U}, {\boldsymbol \Theta} )$  w.r.t the $\Sigma_k$ is computed as follows:
 \begin{equation}
   \frac{\partial  {\cal L}_1({\bf U}, {\boldsymbol \Theta} ) }{\partial \Sigma_k}  =-\frac{1}{2}\sum_{n=1}^{N} \gamma_k({\bf z}_n) \bigg[ \Sigma_{k}^{-1}-\Sigma_{k}^{-1} (\mathbf{z}_n -{\boldsymbol{\mu}_k})(\mathbf{z}_n -{\boldsymbol{\mu}_k})^{T} \Sigma_{k}^{-1} \bigg].
 \end{equation}
When we use the above gradients to update Gaussian covariance matrices in SGD, we have to impose additional constraints to ensure all covariance matrices are positive semidefinite. However, if we adopt diagonal covariance matrices for all Gaussian components, these constraints can be implemented in a fairly simple way. 

Finally, the partial derivative of ${\cal L}_1 ({\bf U}, {\boldsymbol \Theta} )$   w.r.t the projection matrix $\mathbf{U}$ is computed as:
\begin{equation}
  \frac{\partial {\cal L}_1({\bf U}, {\boldsymbol \Theta} ) }{\partial \mathbf{U}}  =  \sum_{n=1}^{N}\sum_{k=1}^{K} \gamma_k({\bf z}_n)\cdot \Sigma_{k}^{-1}(\boldsymbol{\mu}_k-\mathbf{z}_n) \mathbf{x}_n^{T}.
  \end{equation}

\subsection{Computing ${\cal L}_1$ for movMFs}
\label{subsec-MLE-HOPE-vMF}

Similarly, we derive all partial derivatives of ${\cal L}_1({\bf U}, {\boldsymbol \Theta} )$ for a mixture of vMFs (movMFs). In this case, given a mini-batch of training samples, 
${\bf X} = \{ {\bf x}_n \; | \; n=1, \cdots, N \}$,
the log-likelihood function of the HOPE model with movMFs can be expressed as follows:
\begin{equation} \label{moVMF_HOPE}
{\cal L}_1({\bf U}, {\boldsymbol \Theta} ) =	
\sum\limits_{n = 1}^N \;\; \ln \left[  \sum\limits_{k = 1}^K 
\; \pi_k \cdot {\cal C}_M (|{\boldsymbol \mu}_k|) \cdot e^{ {\bf z}_n \cdot {\boldsymbol \mu}_k } \right]
\end{equation}
where each ${\bf z}_n$ must be normalized to be of unit length \footnote{In practice, we usually normalize all original data, ${\bf x}_n$, to be of unit length: $|{\bf x}_n|=1$, prior to the HOPE model. In this case, as long as $M$ is properly chosen (e.g., to be large enough), the projection matrix ${\bf U}$ is always learned to extract from ${\bf x}_n$ as much energy as possible. Therefore, this normalization step may be skipped because the norm of the projected ${\bf z}_n$ is always very close to one even without normalization in this stage, i.e., $|{\bf z}_n| = |\tilde {\bf z}_n | \approx 1$.} as required by the vMF distribution as:

\begin{equation} \label{eq-HOPE-unit-normalization}
\tilde {\bf z}_n = {\bf U}{\bf x}_n,\quad {\bf z}_n = \frac{{\tilde {\bf z}_n}}{{| \tilde{\bf z}_n |}}.
\end{equation}

Similar to the HOPE models with GMMs, we first define an occupancy statistic for $k$-th vMF component as:
\begin{equation}\label{eq-vMF-occupancy}
\gamma_k({\bf z}_n)=\frac{\pi_k\cdot {\cal C}_{M} (|{\boldsymbol \mu}_k|) \cdot e^{{\bf z}_n \cdot {\boldsymbol \mu}_k}}{\sum_{j=1}^{K}\pi_j\cdot {\cal C}_{M} (|{\boldsymbol \mu}_j|) \cdot e^{{\bf z}_n \cdot {\boldsymbol \mu}_j}}.
\end{equation}

In a similar way, we can derive the partial derivatives of ${\cal L}_1({\bf U}, {\boldsymbol \Theta})$ with respect to $\pi_k, \boldsymbol{\mu}_k$ and $\mathbf{U}$ as follows:
\begin{equation}
\frac{\partial {\cal L}_1 ({\bf U}, {\boldsymbol \Theta})}{\partial \pi_k} = \sum_{n=1}^{N} 
\frac{\gamma_k({\bf z}_n)}{\pi_k}
\end{equation}

\begin{equation} \label{eq-vMF-mu}
\frac{{\partial {\cal L}_1({\bf U}, {\boldsymbol \Theta})}}{{\partial {\boldsymbol \mu _k}}} = \sum_{n=1}^{N} \;\; \gamma_k({\bf z}_n)\cdot \bigg[ {\bf z}_n - \frac{\boldsymbol{\mu}_k}{|{\boldsymbol \mu}_k|}\cdot \frac{I_{M/2}(|{\boldsymbol \mu}_k|)}{I_{M/2-1}(|{\boldsymbol \mu}_k|)} \bigg]
\end{equation} 

 \begin{equation}
 \frac{\partial {\cal L}_1 ({\bf U}, {\boldsymbol \Theta} )}{\partial \mathbf{U}}  = \sum_{n=1}^{N} \;\; \sum_{k=1}^{K} \;\; \frac{\gamma_k({\bf z}_n)}{|\tilde{\bf z}_n|}\cdot({\bf I} - {\bf z}_n{\bf z}_n^T) \boldsymbol{\mu}_k {\bf x}_n^{T}
 \end{equation}

Refer to Appendix \ref{appendix_derivatives_movMF} for all details on how to derive the above derivatives for the movMF distributions. Moreover, when movMFs are used, we need some special treatments to compute the Bessel functions in vMFs, i.e, $I_v(\cdot)$, as shown in eqs.(\ref{eq-vMF-occupancy}) and (\ref{eq-vMF-mu}). In this work, we adopt the numerical method in \cite{Abramowitz_1964} to approximate the Bessel functions, refer to the Appendix \ref{appendix-numerical-Bessel} for the numerical details on this. 
 
\subsection{The SGD-based Learning Algorithm}

Because all mixture weights, $\pi_k$ ($k=1, \cdots K$), and all row vectors, 
${\bf u}_i$ ($i=1, \cdots, M$) of the projection matrix satisfy the constraints: $\sum_k^K \pi_k =1$ and 
$ |{\bf u}_j| = 1$ $ (\forall j) $. During the SGD learning process, 
  $\pi_k$
and  ${\bf u}_i$ must be normalized  after each update as follows: 

\begin{equation}
\pi_k \leftarrow \frac{ \pi_k } { \sum\nolimits_j  \pi_j }
\end{equation}

\begin{equation}
{\bf u}_i \leftarrow \frac{ {\bf u}_i }  { |{\bf u}_i |}.
\end{equation}

\begin{algorithm}[tb]  
 \caption{SGD-based Maximum Likelihood Learning Algorithm for HOPE}
\begin{algorithmic} 
\label{algMLEHOPESGD}
   \STATE randomly initialize ${\bf u}_i$ ($i=1, \cdots, M$), $\pi_k$ and ${\boldsymbol \theta}_k$  
 ($k=1, \cdots, K$)
   \FOR{$epoch=1$ {\bfseries to} $T$}
   \FOR{$minibatch$ ${\bf X}$  {\bfseries in} training set}
   \STATE $ {\bf U} \leftarrow  {\bf U} + \epsilon \cdot \left( \frac{\partial {\cal L}_1({\bf U}, {\boldsymbol \Theta})}{\partial {\bf U}} 
+ \frac{\partial {\cal L}_2({\bf U}, \sigma) }{\partial {\bf U}}
- \beta \cdot \frac{\partial {\cal D}({\bf U}) }{ \partial {\bf U} }    \right)    $   
   \STATE ${\boldsymbol \theta}_k \leftarrow  {\boldsymbol \theta}_k + \epsilon \cdot \frac{\partial {\cal L}_1({\bf U}, {\boldsymbol \Theta})}{\partial {\boldsymbol \theta}_k} $ \; ($\forall k$)
   \STATE  $\pi_k  \leftarrow \pi_k + \epsilon \cdot \frac{\partial {\cal L}_1({\bf U}, {\boldsymbol \Theta})}{\partial {\pi_k}}$ \; ($ \forall k $)
  \STATE $\sigma^2 \leftarrow \frac{1}{N (D-M)}\sum_{n =1}^N \; {\bf n}_n^T {\bf n}_n$
  \STATE $\pi_k \leftarrow \frac{ \pi_k } { \sum\nolimits_j  \pi_j }$ \; ($ \forall k $)
 \;\; and \;\; ${\bf u}_i \leftarrow \frac{ {\bf u}_i }  { |{\bf u}_i |}$  \; ($ \forall i $)
   \ENDFOR
   \ENDFOR
\end{algorithmic} 
\end{algorithm}

Finally, we summarize the SGD algorithm to learn the HOPE models based on the maximum likelihood (ML) criterion in Algorithm 1. 

\section{Learning Neural Networks as HOPE}

As described above, the HOPE model may be used as a novel model for high-dimensional data. The HOPE model itself can be efficiently learned unsupervised from unlabelled data  based on the above-mentioned maximum likelihood criterion. 
Moreover, if data labels are available, a variety of discriminative training methods, such as those in \cite{Jiang2010a,Jiang2010b,Jiang2014}, may be used to learn the HOPE model supervised based on some other discriminative learning criteria.  

More interestingly, as we will elucidate here, there exists strong relationship between the HOPE models and neural networks (NNs). First of all, we will show that the HOPE models may be used as a new tool to probe the mechanism why NNs work so well in practice. Under the new HOPE framework, we may explain why NNs can almost universally excel on a variety of data types and how NNs can handle various types of highly-correlated high-dimensional data, which may be quite challenging to many other machine learning models.  Secondly, more importantly, the HOPE framework provides us with some new approaches to learn NNs: (i) Unsupervised learning: the maximum likelihood estimation of HOPE may be directly applied to learn NNs from unlabelled data; (ii) Supervised learning: the HOPE framework can be incorporated into the normal supervised learning of NNs by explicitly imposing some orthogonal constraints in learning. This may improve the learning of NNs and yield better and more compact models. 

\subsection{Linking HOPE to Neural Networks}

\begin{figure}[h]       
    \fbox{\includegraphics[height=2in,width=3in]{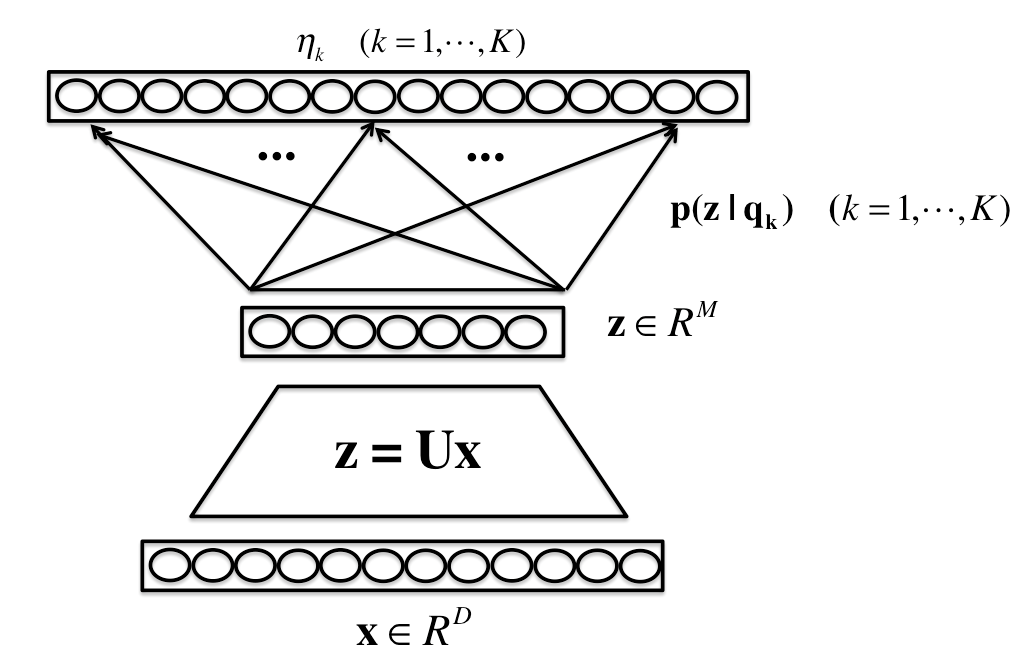} (a)}   
    \hspace{20px}
    \fbox{\includegraphics[height=1.5in,width=2in]{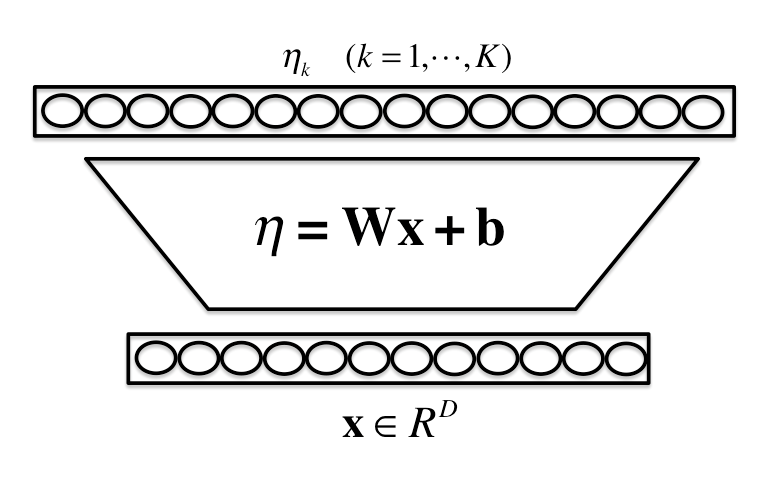} (b)}
    \caption{Illustration of a HOPE model as a layered network structure in (a). It may be equivalently reformulated as a hidden layer in neural nets shown in (b).}
    \label{fg-Hope-onelayer}
\end{figure}

A HOPE model normally consists of two stages: i) a linear orthogonal projection from the raw data space to the latent feature space; ii) a generative model defined as a finite mixture model in the latent feature. As a result, we may depict every HOPE model as a two-layer network: a linear projection layer and a nonlinear model layer, as shown in Figure \ref{fg-Hope-onelayer} (a). The first layer represents the linear orthogonal projection from ${\bf x}$ $({\bf x} \in {\bf R}^D)$ to ${\bf z}$ $({\bf z} \in {\bf R}^M)$:  ${\bf z} =  {\bf U}  {\bf x}$. The second layer represents the underlying finite mixture model in the feature space. Taking movMFs as an example, each node in the model layer represents the log-likelihood contribution from one mixture component as follows: 
\begin{equation}
\phi_k = \ln \left(\pi_k \cdot f_k({\bf z} | {\boldsymbol \theta}_k) \right)
= \ln \pi_k + \ln {\cal C}_{M} (|{\boldsymbol \mu}_k|) + {\bf z} \cdot {\boldsymbol \mu}_k.
\end{equation}

\begin{figure}[h]   
	\begin{center}    
    \includegraphics[width=4in]{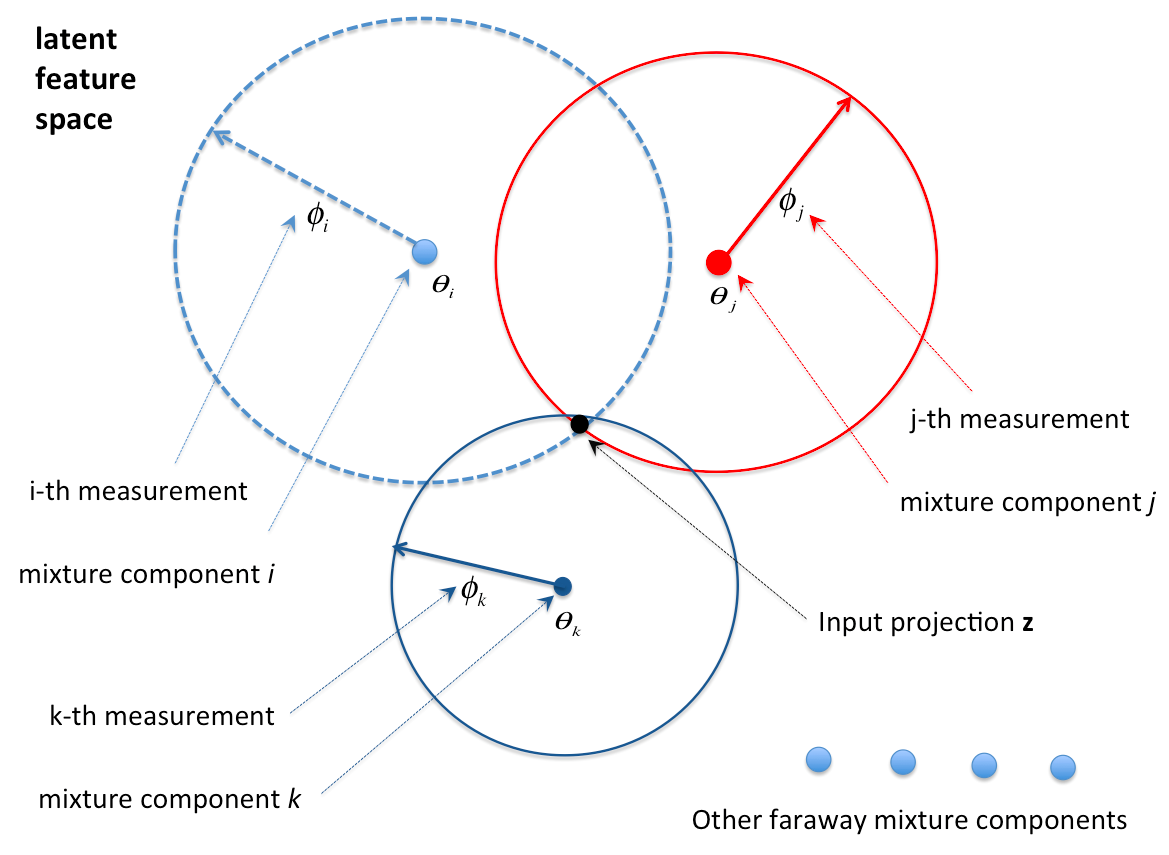}   
    \caption{Illustration of the HOPE features as trilateration in the latent feature space. \label{fg-Hope-trilateration}}
	\end{center}
\end{figure}

Given an input ${\bf x}$ (assuming ${\bf x}$ is projected to ${\bf z}$ in the latent feature space), if we know all $ \phi_k \; (1 \leq k \leq K)$ in the model layer, we can easily compute the log-likelihood value of ${\bf x}$ from the HOPE model in eq.(\ref{eq-movMF-pdf}) as follows:
\begin{equation}
\ln p({\bf z}) = \ln \sum_{k=1}^K  \exp(\phi_k).
\end{equation}
Moreover, all $ \phi_k \; (1 \leq k \leq K)$ may be used as a set of distance measurements to locate the input projection, ${\bf z}$, in the latent space as {\em trilateration}, shown in Figure \ref{fg-Hope-trilateration}. Therefore, all $ \phi_k \; (1 \leq k \leq K)$ may be viewed as a set of features to distinctly represent the original input ${\bf x}$.

Furthermore, we may use a preset threshold $\varepsilon$  to prune the raw measurements, $ \phi_k \; (1 \leq k \leq K)$, as follows:
\begin{equation} \label{eq-HOPE-outputs}
\eta_k = \max(0, \phi_k - \varepsilon)
\end{equation}
to eliminate those small log likelihood values from some faraway mixture components.
Pruning small log-likelihood values as above may result in several benefits. 
Firstly, this pruning operation 
does not affect the total likelihood from the mixture model because it is always dominated by only a few top components. Therefore,  we have:
$$
\ln p({\bf z}) \approx  \varepsilon + \ln \sum_{k=1}^K  \exp(\eta_k).
$$
Secondly, this pruning step does not affect the trilateration problem in Figure \ref{fg-Hope-trilateration}. This is similar to the Global Positioning System (GPS) where the weak signals from faraway satellites are not used for localization. 
More importantly,  the above pruning operation may improve robustness of the features since these small log-likelihood values may become very noisy. 
Note that the above pruning operation is similar to the rectified linear nonlinearity in regular ReLU neural networks. Here we give a more intuitive explanation for the popular ReLU operation under the HOPE framework. 

In this way, as shown in Figure \ref{fg-Hope-onelayer} (a), all rectified log likelihood values in the model layer, i.e., $\eta_k \; (1\leq k \leq K)$, may be viewed as a sensory map to measure each input, ${\bf x}$, in the latent feature space using all mixture components as the probers. Each pixel in the map, i.e., a pruned measurement $\eta_k$, roughly tells the distance between the centroid of a mixture component and the input projection ${\bf z}$ in the latent feature space. Under some condition (e.g., $K \gg M$), the input projection can be precisely located based on these pruned $\eta_k$ values as a trilateration problem in the $M$-dimensional feature space, as shown in Figure \ref{fg-Hope-trilateration}. Therefore, this sensory map may be viewed as a feature representation learned to represent the input ${\bf x}$, which may be fed to a {\em softmax} classifier to form a normal shallow neural network, or to another HOPE model to form a deep neural network. 

Moreover, since the projection layer is linear, it can be mathematically combined with the upper model layer to generate a single layer structure, as shown in Figure \ref{fg-Hope-onelayer} (b).  If movMFs are used in HOPE, it is equivalent to a hidden layer in normal rectified linear (ReLU) neural networks.\footnote{On the other hand, if GMMs are used in HOPE, it can be similarly shown that it is equivalent to a hidden layer in Radial basis function (RBF) networks \cite{RBFNetwork1991}.}  And the weight matrix in the merged layer can be simply derived from the HOPE model parameters, ${\bf U}$ and ${\boldsymbol \Theta}$. It is simple to show that the weight vectors for each hidden node $k$ ($1\leq k \leq K$) in Figure \ref{fg-Hope-onelayer} (b) may be derived  as $${\bf w}_k = {\bf U}^T {\boldsymbol \mu}_k$$ 
and its bias  is computed as $$ b_k = \ln \pi_k + \ln {\cal C}_{M} (|{\boldsymbol \mu}_k|) - \varepsilon.$$

Even though the linear projection layer may be merged with the model layer after all model parameters are learned, however, it may be beneficial to keep them separate during the model learning process. In this way, the model capacity may be controlled by two distinct control parameters: i) $M$ can be selected properly to filter out noise components as in eq.(\ref{eq-HOPE-signals}) to prevent overfitting in learning; ii) $K$ may be chosen independently to ensure the model is complex enough to model very big data sets for more difficult tasks. Moreover, we may enforce the orthogonal constraint, i.e., $ {\bf U} {\bf U}^{T}= {\bf I}$, during the model learning to ensure that all dimensions of ${\bf z}$ are mostly de-correlated in the latent feature space, which may significantly simplify the density estimation in the feature space using a finite mixture model. 

The formulation in Figure \ref{fg-Hope-onelayer} (a) helps to explain the underlying mechanism how neural networks work. Under the HOPE framework, it becomes clear that each hidden layer in neural networks may actually perform two different tasks implicitly, namely feature extraction and data modeling. This may shed some light on why neural nets can directly deal with various types of highly-correlated high-dimensional data \cite{Pan2012} without any explicit dimension reduction and feature de-correlation steps. 

Based on the above discussion, a HOPE model is mathematically equivalent to a hidden layer in neural networks. For example,  each movMF HOPE model can be collapsed into a single weight matrix, same as a hidden layer of ReLU NNs. On the contrary, any weight matrix in ReLU NNs may be decomposed as a product of two matrices as in Figure \ref{fg-Hope-onelayer} (a) as long as $M$ is not less than the rank of the weight matrix. In practice, we may deliberately choose a smaller value for $M$ to regularize the models. 
Under this formulation, it is clear that neural networks may be trained under the HOPE framework.
There are several advantages to learn neural networks under the HOPE framework. First of all, the modelling capacity of neural networks may be explicitly controlled by selecting proper values for $M$ and $K$, each of which is chosen for a different purpose. 
Secondly, we can easily apply the maximum likelihood estimation of HOPE models in section \ref{sec-upsupervised-MLE-HOPE}
to unsupervised or semi-supervised learning to learn NNs from unlabelled data. Thirdly, the useful orthogonal constraints may be incorporated into the normal back-propagation process to learn better NN models in supervised learning as well. 

\subsection{Unsupervised Learning of Neural Networks as HOPE}
\label{subsec-unsupervisedlearning-NN-HOPE}

The maximum likelihood estimation method for HOPE in section \ref{sec-upsupervised-MLE-HOPE} can be used to learn neural networks layer by layer in an unsupervised learning mode. All HOPE model parameters in Figure \ref{fg-Hope-onelayer} (a) are first estimated based on the maximum likelihood criterion as in section \ref{sec-upsupervised-MLE-HOPE}. Next, the two layers in the HOPE model are merged to form a regular NN hidden layer as in Figure \ref{fg-Hope-onelayer} (b). In this case, class labels are not required to learn all network weights and neural networks can be learned from unlabelled data under a theoretically solid framework. This is similar to the Hebbian style learning \cite{EdmundRolls98} but it has a well-founded and converging objective function in learning. As described above, the rectified log-likelihood values from the HOPE model, i.e., $\eta_k \; (1\leq k \leq K)$,  may be viewed as a sensory map using all mixture components as the probers in the latent feature space, which may serve as a good feature representation for the original data.
At the end, a small amount of labelled data may be used to learn a simple classifier, either a {\em softmax} layer or a linear support vector machine (SVM), on the top of the HOPE layers, which takes the sensory map as input for final classification or prediction.

In unsupervised learning, the learned orthogonal projection matrix ${\bf U}$ may be viewed as a generalized PCA, which performs dimension reduction by considering the complex distribution modeled by a finite mixture model in the latent feature space. 
 
\subsection{Supervised Learning of Neural Networks as HOPE}
\label{subsec-supervisedlearning-NN-HOPE}

The HOPE framework can also be applied to the supervised learning of neural networks when data labels are available. 
Let us take ReLU neural networks as example, each hidden layer in a ReLU neural network, as shown in Figure \ref{fg-Hope-onelayer} (b), may be viewed as a HOPE model and thus it can be decomposed as a combination of a projection layer and a model layer, as shown in Figure \ref{fg-Hope-onelayer} (a). In this case, $M$ needs to be chosen properly to avoid overfitting. In other words, each hidden layer in ReLU NNs is represented as two layers during learning, namely a linear projection layer and a nonlinear model layer. 
If data labels are available, as in \cite{Jiang2010b,Jiang2014}, instead of using the maximum likelihood criterion, we may use other discriminative training criteria \cite{Jiang2010a} to form the objective function to learn all network parameters. Let us take the popular minimum cross-entropy error criterion as an example, given a training set of the input data and the class labels, i.e., 
${\bf X} =\{ {\bf x}_t, {l}_t \; | \; 1 \leq t \leq T\}$, we may use the HOPE outputs, i.e., all $\eta_k$ in eq.(\ref{eq-HOPE-outputs}) to form the cross-entropy objective function as follows:
\begin{equation}
{\cal F}_{CE} ({\bf U}, {\boldsymbol \mu}_k, b_k \; | \; {\bf X} ) =
- \sum_{t=1}^T \log\bigg[\frac{\exp \big(\eta_{l_t}({\bf x}_t) \big)}{\sum_{k=1}^K \exp \big(\eta_k({\bf x}_t) \big)} \bigg]
\end{equation}

Obviously, the standard back-propagation algorithm may be used to optimize the above objective function to learn all decomposed HOPE model parameters end-to-end, including ${\bf U}$, $\{ {\boldsymbol \mu}_k, b_k \; | \; 1 \leq k \leq K\}$ from all HOPE layers. 
The only difference is that the orthogonal constraints, as in eq.(\ref{eq-HOPE-orthogonal-constraint-isolated}), must be imposed for all projection layers during training, where the derivatives in eq.(\ref{eq-derivatives-maxtrix-form}) must be incorporated in the standard back-propagation process to update each projection matrix ${\bf U}$ to ensure it is orthonormal. Note that in the supervised learning based on a discriminative training criterion, the unit-length normalization of the data projections in eq.(\ref{eq-HOPE-unit-normalization}) can be relaxed for computational simplicity since this normalization is only important for computing the likelihood for pure probabilistic models. After the learning, each pair of projection and model layers can be merged into a single hidden layer. After merging, the resultant network remains the exactly same network structure as normal ReLU neural networks.  This learning method is related to the well-known low-rank matrix factorization method used  for training deep NNs in speech recognition \cite{SainathLowRank13,JianXueMSR13}. However, since we impose the orthogonal constraints for all projection matrices during the training process, it may lead to more compact models and/or better performance. 

In supervised learning, the learned orthogonal projection matrix ${\bf U}$ may be viewed as a generalized LDA or HDA\cite{Kumar98}, which optimizes the data projection to maximize (or minimize) the underlying discriminative learning criterion. 

\subsection{HOPE for Deep Learning}

\begin{figure}[h]       
    \fbox{\includegraphics[height=2.0in,width=2.6in]{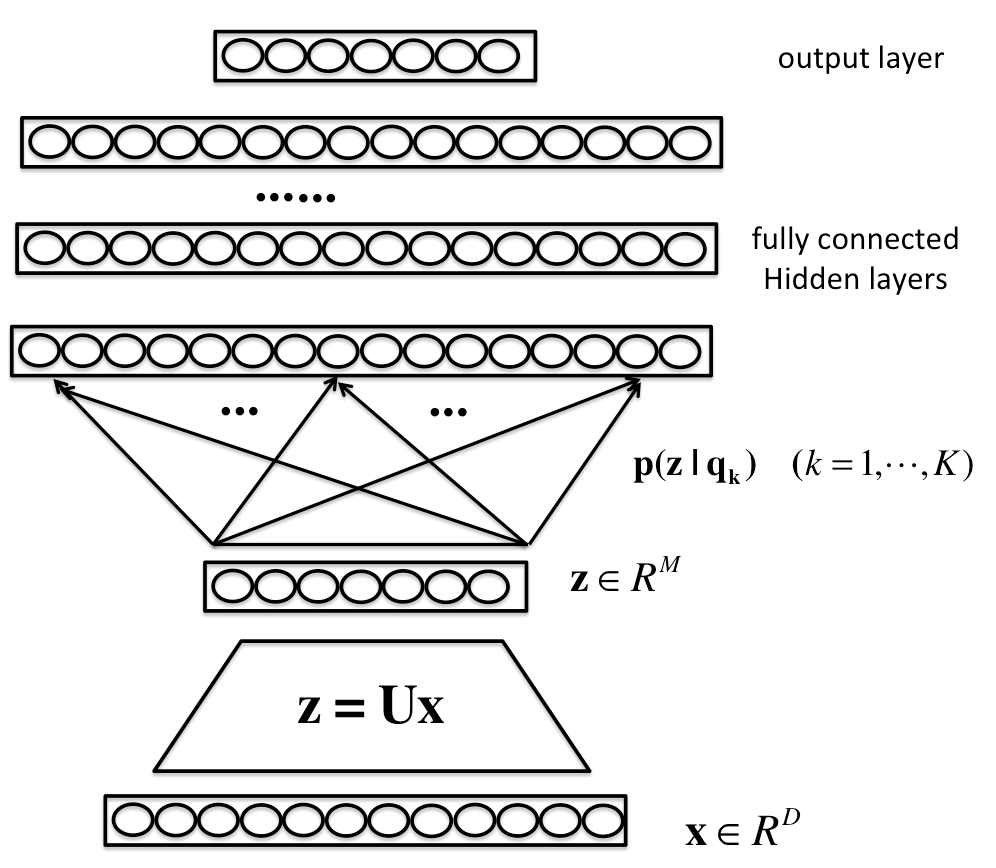} (a)}   
    \hspace{20px}
    \fbox{\includegraphics[height=2.0in,width=2.2in]{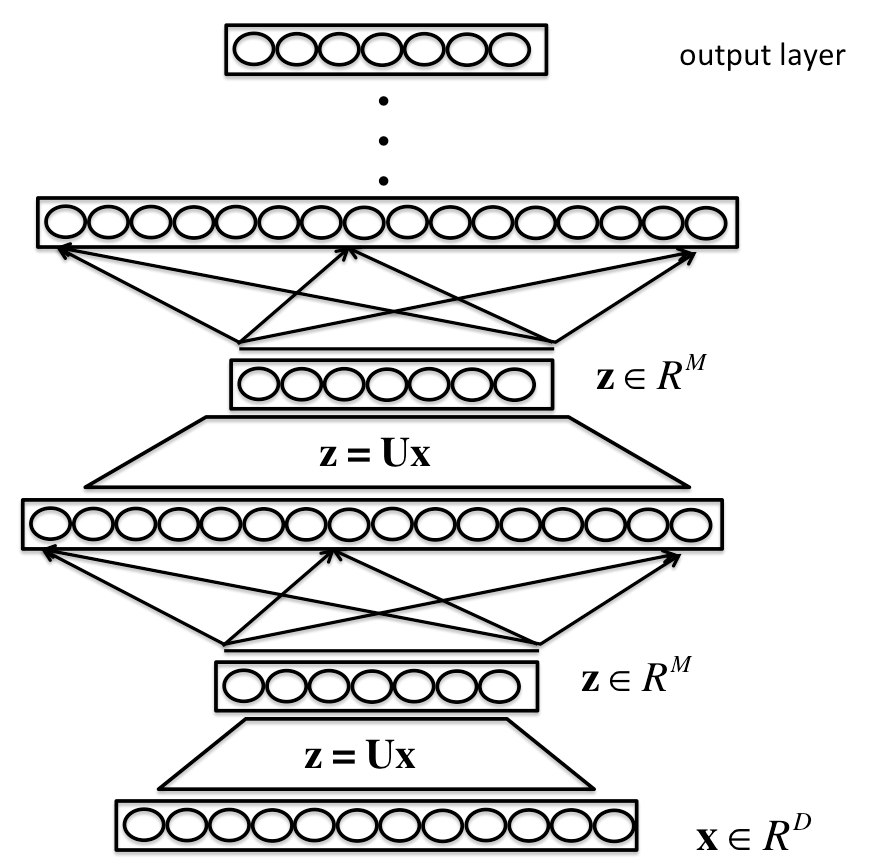} (b)}
    \caption{Two structures to learn deep networks with HOPE: (a) Stacking a DNN on top of one HOPE layer; (b) Stacking multiple HOPE layers. }
    \label{fg-DNN-Hope-Layers}
\end{figure}

As above, the HOPE framework can be used to learn rather strong shallow NN models. 
However, this does not hinder HOPE from building deeper models for deep learning. As shown in Figure \ref{fg-DNN-Hope-Layers}, we may have two different structures to learn very deep neural networks under the HOPE framework. In Figure \ref{fg-DNN-Hope-Layers} (a), one HOPE model is used as the first layer primarily for feature extraction and a deep neural network is concatenated on top of it as a powerful classifier to form a deep structure. 
The deep model in Figure \ref{fg-DNN-Hope-Layers} (a) may be learned in either supervised or semi-unsupervised mode. In semi-unsupervised learning, the HOPE model is learned based on the maximum likelihood estimation and the upper deep NN is learned supervised. Alternatively, if we have enough labelled data, we may jointly learn both HOPE and DNN in a supervised mode.
In Figure \ref{fg-DNN-Hope-Layers} (b), we may even stack multiple HOPE models to form another deep model structure. In this case, each HOPE model generates a sensory feature map in each HOPE layer. Just like all pixels in a normal image, all thresholded log-likelihood values in the sensory feature map are also highly correlated, especially for those mixture components locating relatively close in the feature space. Thus, it makes sense to add another HOPE model on top of it to de-correlate features and perform data modeling at a finer granularity. 
The deep HOPE model structures in Figure \ref{fg-DNN-Hope-Layers}  (b)
can also be learned in a either supervised or unsupervised mode. In unsupervised learning, these HOPE layers are learned layer-wise using the maximum likelihood estimation. In supervised learning, all HOPE layers are learned in back-propagation with orthonormal constraints being imposed to all projection layers. 

\section{Experiments}

In this section, we will investigate the proposed HOPE framework in learning neural networks for several standard image and speech recognition tasks under several different learning conditions: i) unsupervised feature learning;  ii) supervised learning; iii) semi-supervised learning. The examined tasks include the image recognition tasks using the MNIST data set, 
and the speech recognition task using the TIMIT data set.

\subsection{MNIST: Image Recognition}
\label{exp-MNIST}

The MNIST data set \cite{lecun1998gradient} consists of $28\times28$ pixel greyscale images of handwritten digits 0-9, with 60,000 training and 10,000 test examples. In our experiments, 
we first evaluate the performance of unsupervised feature learning using the  HOPE model with movMFs.
Secondly, we investigate the performance of supervised learning of DNNs under the HOPE framework, and further study the effect of the orthogonal constraint in the HOPE framework. Finally, 
we consider a semi-supervised learning scenario with the HOPE models, where all training samples (without labels) are used to learn feature representation unsupervised and then a portion of training data (along with labels) is used to learn post-stage classification models supervised. \footnote{Matlab codes are available at \url{https://wiki.eecs.yorku.ca/lab/MLL/projects:hope:start} for readers to reproduce all MNIST results reported in this paper.} 

\subsubsection{Unsupervised Feature Learning on MNIST}
\label{exp-MNIST-unsupervised-features}

In this experiment, we first randomly extract many small patches from the original unlabelled training images on MNIST. Each patch is of 6-by-6 in dimension, represented as a vector in $\mathbb{R}^{D}$, with $D=36$. In this work, we randomly extract 
400,000 patches in total from the MNIST training set for unsupervised feature learning. 
Moreover,  every patch is normalized by subtracting the mean and being  divided by the standard deviation of its elements.  
 
In the unsupervised feature learning, we follow the same experimental setting  in \cite{Coates2011}, where an unsupervised learning algorithm is used to learn a ``black box'' feature extractor to map each input vector in $\mathbb{R}^{D}$ to 
another $K$-dimension feature vector. In this work, we have examined several different unsupervised learning algorithms for feature learning: (i) kmeans clustering; (ii) spherical kmeans (spkmeans) clustering; (iii) mixture of vMF (movMF), (iv) PCA based dimension reduction plus movMF (PCA-movMF); and (v) the HOPE model with movMFs (HOPE-movMF). As for {\em kmeans} and {\em spkmeans}, the only difference is that different distance measures are used in clustering: {\em kmeans} uses the Euclidean distance while {\em spkmeans} uses the cosine distance. As for the movMF model, we can use the expectation maximization (EM) algorithm for estimation, as described in \cite{Banerjee2005}. In the following, we  briefly summarize the experimental details for these feature extractors. 
 \begin{enumerate}
 	\item {\bf kmeans}: We first apply the k-means clustering method to learn $K$ centroids ${\boldsymbol \mu}_{k}$ from all extracted patch input vectors. For each learned centroid ${\boldsymbol \mu}_{k}$,  we use a soft threshold function to compute each feature as: $f_{k}(x) = \max(0, |{\bf x}-{\boldsymbol \mu}^{k}|-\varepsilon)$, where $\varepsilon$ is a pre-set threshold. In this way, we may generate a $K$-dimension feature vector for each patch input vector.  
 	\item {\bf spkmeans}: As for the spk-means clustering, we need to normalize all input patch vectors to be of unit length before clustering them into $K$ different centroids based on the cosine distance measure.  Given each learned centroid ${\boldsymbol \mu}_{k}$, we can compute each feature as  $f_{k}(x) = \max(0,  {\bf x}^T {\boldsymbol \mu}_{k}-\varepsilon)$.
 	\item {\bf movMF}: We also need to normalize all input patch vectors to be of unit length. We  use the EM algorithm to learn all model parameters $\boldsymbol{\mu}_k$. For each learned centroid ${\boldsymbol \mu}_k$ of the movMF model, we  compute one  feature as $f_{k}(x) = \max \big(0, \ln(\pi_k)+\ln({\cal C}_N(| {\boldsymbol \mu}_{k}|)) + {\bf x}^T {\boldsymbol \mu}_k - \varepsilon \big)$. 
 	\item {\bf PCA-movMF}:  Comparing to movMF, the only difference is that we first use PCA for dimension reduction, reducing all input patch vectors from $\mathbb{R}^{D}$ to $\mathbb{R}^{M}$. Then, we use the same method to estimate an movMF model for the reduced $D$-dimension feature vectors. In this experiment, we set $M=20$ to reserve 99.5\% of the total sum of all eigenvalues in PCA. For each learned vMF model ${\boldsymbol \mu}_k$, we  compute one  feature as $f_{k}(x) = \max \big(0, \ln(\pi_k)+\ln({\cal C}_N(| {\boldsymbol \mu}_{k}|)) + {\bf z}^T {\boldsymbol \mu}_k - \varepsilon \big)$. 
	\item {\bf HOPE-movMF}: We use the maximum likelihood estimation method described in section \ref{sec-upsupervised-MLE-HOPE} to learn a HOPE model with movMFs. For each learned vMF component,
we can compute one  feature as $f_{k}(x) = \max(0, \ln(\pi_k)+\ln({\cal C}_M(| {\boldsymbol \mu}_{k} |)) + \tilde{\bf z}\cdot {\boldsymbol \mu}_k - \varepsilon)$. Similar to PCA-movMF, we also set $M=20$ here. 
 \end{enumerate}
Furthermore, since HOPE-movMF is learned using SGD,  we need to tune some hyper-parameters for HOPE, such as learning rate, mini-batch size, $\beta$ and $\sigma^2$. In this work, the learning rate is set to 0.002, minibatch size is set to 100, we set $\beta=1.0$, and the noise variance is manually set to $\sigma^2=0.1$ for convenience. 

After learning the above models, they are used as feature extractors. We use the same method as described in \cite{Coates2011} to 
generate a feature representation for each MNIST image, where  each feature extractor is convolving over an MNIST image to obtain the feature representations for all local patches in the image. Next, we split the image into four equally-sized quadrants and 
the feature vectors in each quadrant are all summed up to pool as one feature vector. In this way,  we can get a $4K$-dimensional feature vector for each MNIST image, where $K$ is the number of all learned features for each patch. Finally, we use these pooled $4K$-dimension feature vectors for all training images, along with the labels, to estimate a simple linear SVM as a post-stage classifier for image classification.  The experimental results are shown in Table \ref{MNIST_USL}. 
We can see that {\em spkmeans} and {\em movMF} can achieve much better performance than {\em kmeans}. The PCA-based dimension reduction leads to further performance gain. Finally, the jointly trained HOME model with movMFs yields the best performance, e.g., $0.64\%$ in classification error rate when $K=1200$. 
This is a very strong performance for unsupervised feature learning on MNIST.


\begin{table}
	\centering
	\caption{Classification error rates (in \%) on the MNIST test set using supervised learned linear SVMs as classifiers and unsupervised learned features (4K-dimension) from different models. }
	\begin{tabular}{|c|c|c|c|c|}
		\hline   model / K=     &  400   & 800   & 1200  & 1600  \\ 
		\hline  kmeans     & 1.41  & 1.31  & 1.16  & 1.13 \\ 
		\hline  spkmeans & 1.09  & 0.90  &  0.86 & 0.81 \\ 
	    \hline  movMF      & 0.89  & 0.82  & 0.81   & 0.84  \\ 
		\hline  PCA-movMF& 0.87 & 0.75  & 0.73 & 0.74  \\ 
		\hline\hline  HOPE-movMF& 0.76 & 0.71  & {\bf 0.64}  & 0.67 \\ 
		\hline 
	\end{tabular} 
	\label{MNIST_USL}
\end{table}

\subsubsection{Supervised Learning of Neural Networks as HOPE on MNIST}

In this experiment, we use the MNIST data set to examine the supervised learning of rectified linear (ReLU) neural networks under the HOPE framework, as discussed in section \ref{subsec-supervisedlearning-NN-HOPE}.

Here we follow the normalized initialization in \cite{glorot2010understanding} to randomly initialize all NN weights, without using pre-training.  We adopt a small modification to the method  in \cite{glorot2010understanding} by adding a factor to control the dynamic range of initial weights as follows:
\begin{equation}
W_i \sim \bigg[-\gamma\cdot\frac{\sqrt{6}}{\sqrt{n_{i}+n_{i+1}}},\gamma \cdot\frac{\sqrt{6}}{\sqrt{n_{i}+n_{i+1}}} \bigg]
\end{equation}
where $n_{i}$ denotes the number of units in the $i$-th layer.
For the ReLU units, due to the unbounded behavior of the activation function, activations of ReLU units might grow unbounded. To handle this numerical problem, we shrink the dynamic range of initial weights by using a small factor $\gamma \; (\gamma=0.5)$, which is equivalent to scaling the activations. 

We use SGD to train  ReLU neural networks using the  following learning schedule:
\begin{equation}
\epsilon^{t}= \epsilon_{0}\cdot \alpha^{t}
\end{equation}

\begin{equation}
m^{t}=
\begin{cases}
\frac{t}{T} m_{f} + (1 - \frac{t}{T}) m_{i} & \text{$t<T$} \\
m_{f} & \text{$ t\geq T $}
\end{cases}
\end{equation}
where 
$\epsilon^{t}$ and $m^{t}$ denote the learning rate and momentum for the $t$-th epoch, 
and we set all control parameters as $ m_{i}=0.5,m_{f}=0.99, \alpha=0.998$.  We totally run $T=50$ epochs for learning without dropout and run $T=500$ epochs for learning with dropout. Moreover, the weight decay is used here and it is set to 0.00001. 
Furthermore, for the HOPE model, the control parameter for the orthogonal constraint, $\beta$,  is set to 0.01 in all experiments.
In this work, we do not use any data augmentation method. 

\begin{figure}
	\centering
	\includegraphics[width=1\linewidth]{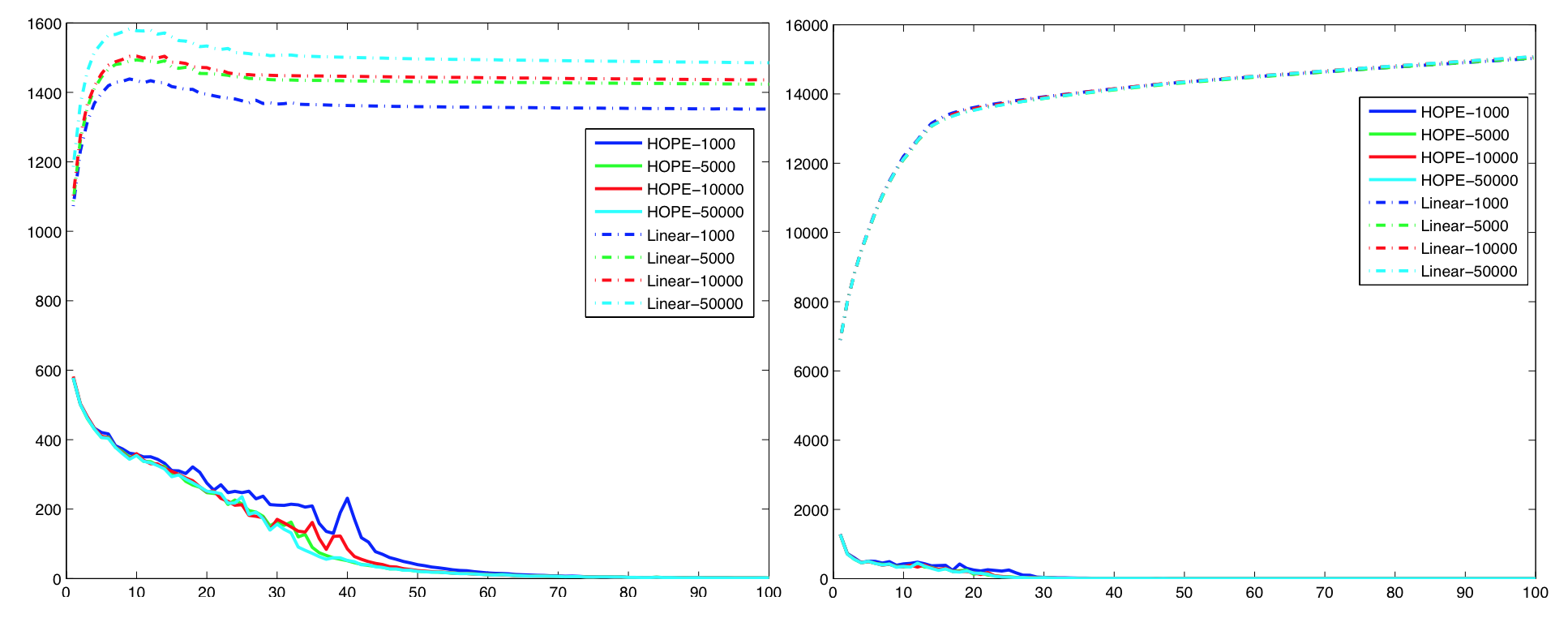}
	\caption{The learning curves of the total sum of correlation coefficients  of the linear projection and orthogonal HOPE projection matrices, respectively. left: $M$=200, $K$=1k,5k,10k,50k; right: $M$=400, $K$=1k,5k,10k,50k.}
	\label{fig:Corr_curve}
\end{figure}

\begin{figure}
\centering
\includegraphics[width=1\linewidth]{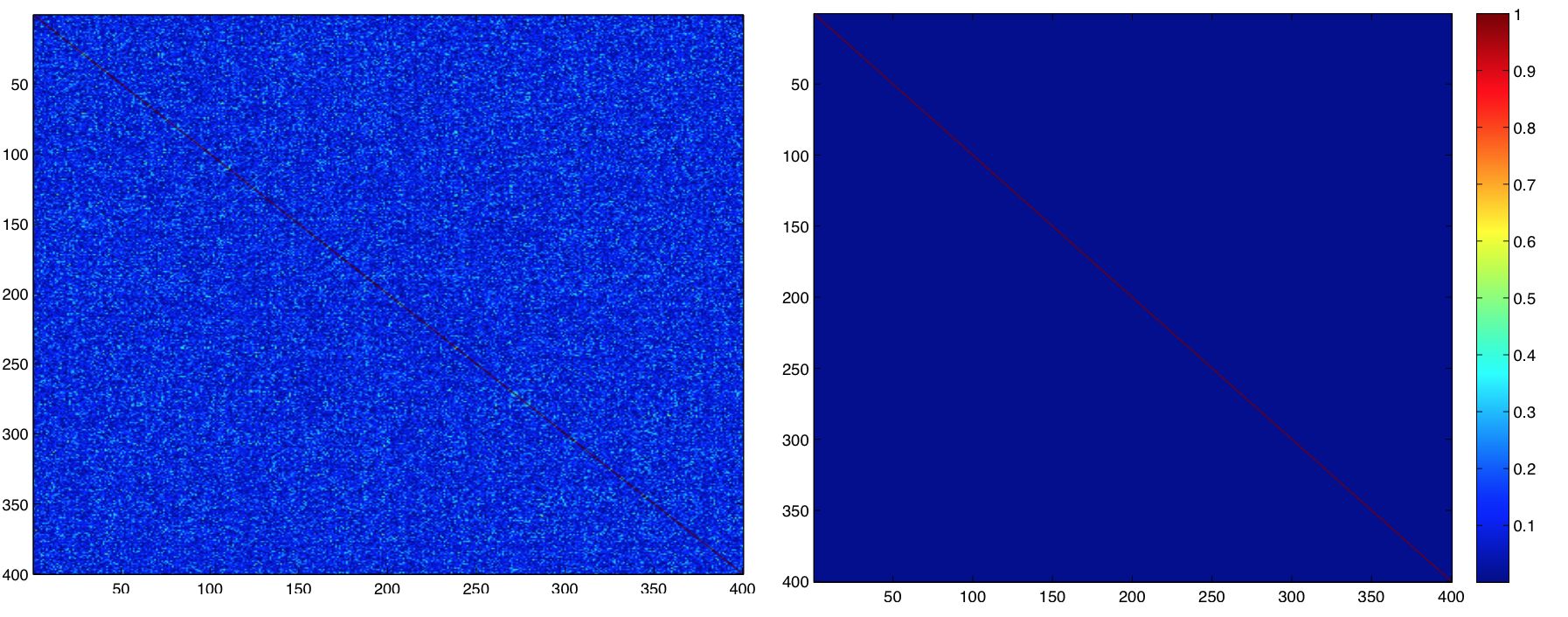}
\caption{The correlation coefficients of the linear projection matrix (left) and the orthogonal HOPE projection matrix (right) are shown as two images. Here $M=400$ and $K=1000$. }
\label{fig:ImageShow_Corr}
\end{figure}

Under the HOPE framework, we decompose each ReLU hidden layer into two layers as in Figure \ref{fg-Hope-onelayer} (a). In this experiment, we first examine the supervised learning of NNs with or without imposing the orthogonal constraints to all projection layers.  
Firstly, we investigate the performance of a neural network containing only a single hidden layer, decomposed into a pair of a linear projection layer and a nonlinear model layer.  Here we evaluate neural networks with a different number of hidden units ($K$) and a varying size of the projection layer ($M$). From the experimental results shown in Table. \ref{MNIST_SL_1}, we can see that 
the HOPE-trained NN can achieve much better performance than the baseline NN, especially when smaller values are used for $M$. This supports that the projection layer may eliminate residual noises in data to avoid over-fitting when $M$ is properly set. 
However, after we relax the orthogonal constraint in the HOPE model, as shown in Table \ref{MNIST_SL_1}, the performance of the models using  only linear projection layers gets much worse than those of the HOPE models as well as that of the baseline NN. 
These results verify that orthogonal projection layers are critical in the HOPE models.
Furthermore, in Figure. \ref{fig:Corr_curve}, we have plotted the learning curves of the total sum of all correlation coefficients among all row vectors in the learned projection matrix ${\bf U}$, i.e., $\sum_{i \neq j}\; \frac{|{\bf u}_i \cdot {\bf u}_j|}{|{\bf u}_i| |{\bf u}_j|}$. 
We can see that all projection vectors tend to get strongly correlated (especially when $M$ is large) in the linear projection matrix as the learning proceeds. On the other hand, the orthogonal constraint can effectively de-correlate all the projection vectors. 
Moreover, we show  all correlation coefficients, i.e., $\frac{|{\bf u}_i \cdot {\bf u}_j|}{|{\bf u}_i| |{\bf u}_j|}$,  of the linear projection matrix and the HOPE orthogonal projection matrix as two images in Figure \ref{fig:ImageShow_Corr}, which clearly shows that the linear projection matrix has many strongly correlated dimensions and the HOPE projection matrix is (as expected) orthogonal . 

\begin{table}
	\centering
	\caption{Classification error rates (in \%) on the MNIST test set using a shallow NN with one hidden layer of different hidden units. Neither dropout nor data augmentation is used here for quick turnaround. Two numbers in brackets, $[M,K]$, indicate a HOPE model with the orthogonal constraint in the projection, and two number in parentheses, $(M,K)$, indicate the same model structure without imposing the orthogonal constraint in the projection.}
	\begin{tabular}{|c|c|c|c|c|c|} \hline
		 Net Architecture / K= &  1k& 2k &5k  &10k  &50k  \\ \hline
		 Baseline:  784-K-10 & 1.49  & 1.35 & 1.28 & 1.30 & 1.32 \\ \hline \hline
		 HOPE1: 784-[100-K]-10 & 1.18 & 1.20 & 1.17 & 1.18 & 1.19 \\ \hline
		 HOPE2: 784-[200-K]-10& 1.21 & 1.20 & 1.17 & 1.19 &  1.18\\ \hline
		 HOPE3: 784-[400-K]-10& 1.19 & 1.23 &  1.25& 1.25 & 1.25 \\ \hline \hline
		 Linear1: 784-(100-K)-10 & 1.45 & 1.49 & 1.43 & 1.45 & 1.48 \\ \hline
		 Linear2: 784-(200-K)-10 & 1.52 & 1.50 &  1.54& 1.55 & 1.54 \\ \hline
		 Linear3: 784-(400-K)-10& 1.53 & 1.52 & 1.49 & 1.52 & 1.49 \\ \hline
		\end{tabular} 
		\label{MNIST_SL_1}
\end{table}

As the MNIST training set is very small, we further use the dropout technique 
in \cite{Hinton2012Dropout} to improve the model learning on the MNIST task. 
In this experiment, the visible dropout probability is set to 20\% and the hidden layer dropout probability is set to 30\%. 
In Table \ref{MNIST_SL_2}, we compare a 1-hidden-layer shallow NN with two HOPE models ($M$=200,400). The results show that the HOPE framework can significantly improve supervised learning of NNs. Under the HOPE framework, we can train very simple shallow neural networks from scratch, which can yield comparable performance as deep models. For example, on the MNIST task, as shown in Table \ref{MNIST_SL_2}, we may achieve 0.85\% in classification error rate using a shallow neural network (with only one hidden layer of 2000 nodes) trained under the HOPE framework.
Furthermore, we consider to build deeper models (two-hidden-layer NNs) under the HOPE framework. Using the two different structures in Figure \ref{fg-DNN-Hope-Layers}, we can further improve the classification error rate to $0.81\%$, as shown in Table. \ref{MNIST_SL_3}. To the best of our knowledge, this is one of the best results reported on MNIST without using CNNs and data augmentation. 

\begin{table}{}
	\centering
	\caption{Comparison of classification error rates (in \%) on the MNIST test set between a shallow NN and two HOPE-trained NNs with the same model structure.  Dropout is used in this experiment.}
	\begin{tabular}{|c|c|c|c|c|c|} \hline
		 Net Architecture / K= &  1k& 2k &5k \\ \hline 
		 NN baseline: 784-K-10& 1.05 & 1.01 & 1.01 \\ \hline \hline
		 HOPE1: 784-[200-K]-10 & 0.99 & {\bf 0.85} & 0.89 \\\hline 
		 HOPE2: 784-[400-K]-10 & 0.86 & 0.86 & {\bf 0.85} \\\hline
	\end{tabular} 
	\label{MNIST_SL_2}
\end{table}

\begin{table}
	\centering
	\caption{Comparison of classification error rates (in \%) on the MNIST test set between a 2-hidden-layer DNN and two HOPE-trained DNNs with similar model structures, with or without using dropout. }
	\begin{tabular}{|c|c|c|c|c|c|c|} \hline
      model & Net Architecture &  without dropout & with dropout \\\hline
	  DNN baseline &  784-1200-1200-10 & 1.25 & 0.92 \\\hline \hline
	  HOPE + NN & 784-[400-1200]-1200-10 & 0.99 & 0.82\\\hline 
	  HOPE*2 & 784-[400-1200]-[400-1200]-10 & 0.97 & {\bf 0.81} \\\hline 	
	\end{tabular} 
	\label{MNIST_SL_3}
\end{table}

\subsubsection{Semi-supervised Learning on MNIST}

In this experiment, we combine the unsupervised feature learning with supervised model learning and examine the classification performance when only limited labelled data is available. Here we also list the results using convolutional deep belief networks (CDBN) in \cite{Lee2009CDBN} as a baseline system. In our experiments, we use the raw pixel features and unsupervised learned (USL) features in section \ref{exp-MNIST-unsupervised-features}. As example, we choose the unsupervised learned features from the HOPE-movMF model ($K=800$) in Table \ref{MNIST_USL}\footnote{For the HOPE-movMF model with $K=800$, there are 115 empty clusters. Thus, the unsupervised learned features are of 2740 in dimension.}.
Next, we concatenate the features to a post-stage classifier, which is supervised trained using only a portion of the training data, ranging from 1000 to 60000 (all). We consider many different types of classifiers here, including linear SVM, regular DNNs and HOPE-trained DNNs. Note that all classifiers are trained separately from the feature learning. 
All results are summarized in  Table \ref{MNIST_Semi_SL}. It shows that we can achieve the best performance when we combine the HOPE-trained USL features with HOPE-trained post-stage classifiers. The gains are quite dramatic no matter how much labelled data is used. For example, when only $5000$ labelled samples are used, our method can achieve 0.90\% in error rate, which significantly outperforms all other methods including CDBN in \cite{Lee2009CDBN}. At last, as we use all training data for the HOPE model, we can achieve 0.40\% in error rate. To the best of our knowledge, this is one of the best results reported on MNIST without using data augmentation. Furthermore, our best system uses a quite simple model structure, consisting of a HOPE-trained feature extraction layer of 800 nodes and a HOPE-trained NN of two hidden layers (1200 node in each layer), which is much smaller and simpler than those top-performing systems on MNIST.

\begin{table}
	\centering
	\caption{Classification error rates (in \%) on the MNIST test set using raw pixel features or unsupervised learned features, along with different post-stage classifiers trained separately by limited labeled data. Here USL denotes the unsupervised learned features from HOPE-movMF ($K=800$). All used classifiers include: DNN1 (784-1200-1200-10);  HOPE-DNN1 (784-[400-1200]-10); HOPE-DNN2 (784-[400-1200]-1200-10); DNN2 (2740-1200-1200-10); HOPE-DNN3 (2740-[400-1200]-10); HOPE-DNN4 (784-[400-1200]-1200-10).}
	\begin{tabular}{|c|c|c|c|c|c|c|} \hline
		 labeled training samples & 1000 & 2000  & 5000 & 10000  & 20000  & 60000(ALL)  \\ \hline
		 CDBN\cite{Lee2009CDBN} &  2.62 & 2.13  & 1.59  & - &  - & 0.82 \\ \hline \hline
		 Raw feature+DNN1 & 8.32 & 4.71 & 3.2  & 2.15 & 1.48 & 0.92  \\ \hline
		 Raw feature+HOPE-DNN1 &  8.22& 4.53 & 2.92 & 2.04 & 1.34  & 0.86  \\ \hline
		 Raw feature+HOPE-DNN2 & 7.21 & 4.02 & 2.60 & 1.83 & 1.30  & 0.82 \\ \hline \hline
		 USL feature+linear SVM & 2.91 & 2.38 & 1.47 & 1.13 & 0.90 & 0.71 \\ \hline
		 USL feature+DNN2 & 2.83 & 1.99 & 1.03 & 0.88  & 0.70 & 0.43 \\ \hline
		 USL feature+HOPE-DNN3 & 2.50  & 1.78 & 0.95 & 0.87  & 0.67 & 0.42 \\ \hline
		 USL feature+HOPE-DNN4 & 2.46 & 1.70 & 0.90 & 0.79 & 0.66 & {\bf 0.40} \\ \hline
	\end{tabular} 
	\label{MNIST_Semi_SL}
\end{table}

\subsection{TIMIT: Speech Recognition}
\label{exp-TIMIT}

In this experiment, we examine the supervised learning of shallow and deep neural networks under the HOPE framework 
for a standard speech recognition task using the TIMIT data set. The HOPE-based supervised learning method is compared with  the regular back-propagation training method. We use the minimum cross-entropy learning criterion here.

The TIMIT speech corpus consists of a training set of 462 speakers, a separate development set of 50 speakers for cross-validation, and a core test set of 24 speakers. All results are reported on the 24-speaker core test set. 
The speech waveform data is analyzed using a 25-ms Hamming window with a 10-ms fixed frame rate. The speech feature
vector is generated by a Fourier-transform-based filter-banks that include 40 coefficients distributed on the Mel scale and
energy, along with their first and second temporal derivatives. This leads to a 123-dimension feature vector per speech frame.
We further concatenate 15 consecutive frames within a long context window of (7+1+7) to feed to the models, as 1845-dimension input vectors \cite{XueShaofei2014}.  
All speech data are normalized 
by subtracting the mean of the training set and being divided by the standard deviation of the training set on each dimension 
so that all input vectors have zero mean and unit variance.
We use 183 target class labels (i.e., 3 states for each
of the 61 phones) for the DNN training. After decoding, these 61
phone classes are mapped to a set of 39 classes for the final
scoring as in \cite{TIMIT-89}. In our experiments, a bi-gram language
model at phone level, estimated from all transcripts in the
training set, is used for speech recognition.

We first train ReLUs based shallow and deep neural networks as our baseline systems. The networks are trained using the  back-propagation algorithm, with a mini-batch size of 100. The initial learn rate is set to 0.004 and it is kept unchanged if 
the error rate on the development set is still decreasing. Afterwards, the learning rate is halved after every epoch, and the whole training procedure is stopped when the error reduction on the development set is less than 0.1\% in two consecutive iterations. In our experiments, we also use momentum and weight decay, which are set to 0.9 and 0.0001, respectively.  When we use the mini-batch SGD to train neural networks under the HOPE framework, the control parameter for the orthogonal constraints, i.e. $\beta$, is set to be 0.01. 

In our experiments, we compare the standard NNs with the HOPE-trained NNs for two network architectures, one shallow network with one hidden layer of 10240 hidden nodes and one deep network with 3 hidden layers of 2048 nodes. The performance comparison between them is shown in Table. \ref{exp-TIMIT-results}. Our results are comparable with \cite{XueShaofei2014} and another recent work \cite{NIPS2014_DNN_no_need_deep}, using deep neural networks on TIMIT.
From the results, we can see that the HOPE-trained NNs can consistently outperform the regular NNs by an about 
0.8\% absolute reduction in phone recognition error rates. Moreover, the HOPE-trained neural networks are much smaller than their counterpart DNNs in number of model parameters if the HOPE layers are not merged. After merging, they have the exactly the same model structure as their counterpart NNs. 

\begin{table}{}
\centering
\caption{ Supervised learning of neural networks on TIMIT with and without using the HOPE framework. The two numbers in a bracket, $[M,K]$, indicate a HOPE model with M-dimension features and K mixture components. FACC: frame classification accuracy from neural networks. PER: phone error rate in speech recognition. }
\begin{tabular}{|c|c|c|c|}	\hline  
model & Net Architecture & FACC (\%) & PER (\%) \\  \hline \hline
NN  &  1845-10240-183 & 61.45  & 23.85 \\ \hline
HOPE-NN & 1845-[256-10240]-183 & 62.11 &  {\bf 23.04} \\ \hline \hline
DNN & 1845-3*2048-183& 63.13 &  22.37 \\ \hline 
HOPE-DNN & 1845-[512-2048]-2*2048-183& 63.55 & {\bf 21.59} \\ \hline 
\end{tabular} 
\label{exp-TIMIT-results}
\end{table}


\section{Final Remarks}

In this paper, we have proposed a novel model, called hybrid orthogonal projection and estimation (HOPE), for high-dimensional data. The HOPE model combines feature extraction and data modeling under a unified generative modeling framework so that both feature extractor and data model can be jointly learned either supervised or unsupervised. More interestingly, we have shown that the HOPE models are closely related to neural networks in a way that each hidden layer in NNs can be reformulated as a HOPE model. Therefore, the proposed HOPE related learning algorithms can be easily applied to perform either supervised or unsupervised learning for neural networks. We have evaluated the proposed HOPE models in learning NNs on several standard tasks, including image recognition on MNIST and speech recognition on TIMIT. Experimental results have strongly supported that the HOPE models can provide a very effective unsupervised learning method for NNs. Meanwhile, the supervised learning of NNs can also be conducted under the HOPE framework, which normally yields better performance and more compact models. 

We are currently investigating the HOPE model to learn convolution neural networks (CNNs) for more challenging image recognition tasks, such as CIFAR and ImageNet. At the same time, we are also examining the HOPE-based unsupervised learning for various natural language processing (NLP) tasks. These results will be reported in the future.

\bibliography{HOPE}

\begin{thebibliography}{10}

\bibitem{Abramowitz_1964}
M.~Abramowitz and I.~A. Stegun.
\newblock {\em Handbook of mathematical functions with formulas, graphs, and
  mathematical tables}.
\newblock Washington : U.S. Govt. Print., 2006.

\bibitem{NIPS2014_DNN_no_need_deep}
J.~Ba and R.~Caruana.
\newblock Do deep nets really need to be deep?
\newblock In {\em Advances in Neural Information Processing Systems (NIPS) 27},
  pages 2654--2662. Curran Associates, Inc., 2014.

\bibitem{Banerjee2005}
A.~Banerjee, I.~S. Dhillon, and J.~Ghosh.
\newblock Clustering on the unit hypersphere using von mises-fisher
  distributions.
\newblock In {\em Journal of Machine Learning Research}, pages 1345--1382,
  2005.

\bibitem{Bao2013}
Y.~Bao, H.~Jiang, L.~Dai, and C.~Liu.
\newblock Incoherent training of deep neural networks to de-correlate
  bottleneck features for speech recognition.
\newblock In {\em Proc. of IEEE International Conference on Acoustics, Speech,
  and Signal Processing (ICASSP)}, 2013.

\bibitem{Bengio07greedylayer-wise}
Y.~Bengio, P.~Lamblin, D.~Popovici, and H.~Larochelle.
\newblock Greedy layer-wise training of deep networks.
\newblock In {\em Advances in Neural Information Processing Systems (NIPS) 19}.
  MIT Press, 2007.

\bibitem{Bishop_PRML_06}
C.~M. Bishop.
\newblock {\em Pattern Recognition and Machine Learning}.
\newblock Springer, 2006.

\bibitem{Bottou04}
L.~Bottou.
\newblock Stochastic learning.
\newblock In {\em Advanced Lectures on Machine Learning (edited by O. Bousquet
  and U. von Luxburg)}, pages 146--168. Springer Verlag, 2004.

\bibitem{Bottou11}
L.~Bottou and O.~Bousquet.
\newblock The tradeoffs of large scale learning.
\newblock In {\em Optimization for Machine Learning (edited by S. Sra, S.
  Nowozin and S. J. Wright)}, pages 351--368. MIT Press, 1964.

\bibitem{Coates2011}
A.~Coates, A.~Y. Ng, and H.~Lee.
\newblock An analysis of single-layer networks in unsupervised feature
  learning.
\newblock In {\em International Conference on Artificial Intelligence and
  Statistics}, pages 215--223, 2011.

\bibitem{glorot2010understanding}
X.~Glorot and Y.~Bengio.
\newblock Understanding the difficulty of training deep feedforward neural
  networks.
\newblock In {\em International Conference on Artificial Intelligence and
  Statistics}, pages 249--256, 2010.

\bibitem{Hinton_1997}
G.~E. Hinton, P.~Dayan, and M.~Revow.
\newblock Modelling the manifolds of images of handwritten digits.
\newblock {\em IEEE Transactions on Neural Networks}, 8(1):65--74, 1997.

\bibitem{Hinton_2006a}
G.~E. Hinton, S.~Osindero, and Y.~W. Teh.
\newblock A fast learning algorithm for deep belief nets.
\newblock {\em Neural computation}, 18:1527–1554, 2006.

\bibitem{Hinton_2006}
G.~E. Hinton and R.~R. Salakhutdinov.
\newblock Reducing the dimensionality of data with neural networks.
\newblock {\em Science}, 313(5786):504--507, 2006.

\bibitem{Hinton2012Dropout}
G.~E. Hinton, N.~Srivastava, A.~Krizhevsky, I.~Sutskever, and R.~R.
  Salakhutdinov.
\newblock Improving neural networks by preventing co-adaptation of feature
  detectors.
\newblock In {\em preprint arXiv 1207.0580}, 2012.

\bibitem{Jiang2010a}
H.~Jiang.
\newblock Discriminative training for automatic speech recognition: A survey.
\newblock {\em Computer and Speech, Language}, 24(4):589--608, 2010.

\bibitem{Jiang2010b}
H.~Jiang and X.~Li.
\newblock Parameter estimation of statistical models using convex optimization:
  An advanced method of discriminative training for speech and language
  processing.
\newblock {\em IEEE Signal Processing Magazine}, 27(3):115--127, 2010.

\bibitem{Jiang2014}
H.~Jiang, Z.~Pan, and P.~Hu.
\newblock Discriminative learning of generative models: large margin
  multinomial mixture models for document classification.
\newblock {\em Pattern Analysis and Applications}, 2014.

\bibitem{Kambhatla_1997}
N.~Kambhatla and T.~K. Leen.
\newblock Dimension reduction by local principal component analysis.
\newblock {\em Neural Computation}, 9(7):1493--1516, 1997.

\bibitem{Kumar98}
N.~Kumar and A.~G. Andreou.
\newblock Heteroscedastic discriminant analysis and reduced rank {HMM}s for
  improved speech recognition.
\newblock {\em Speech Communication}, 26(4):283--297, 1998.

\bibitem{lecun1998gradient}
Y.~LeCun, L.~Bottou, Y.~Bengio, and P.~Haffner.
\newblock Gradient-based learning applied to document recognition.
\newblock {\em Proceedings of the IEEE}, 86(11):2278--2324, 1998.

\bibitem{Lee2009CDBN}
H.~Lee, R.~Grosse, R.~Ranganath, and A.~Ng.
\newblock Convolutional deep belief networks for scalable unsupervised learning
  of hierarchical representations.
\newblock In {\em Proceedings of the 26th Annual International Conference on
  Machine Learning}, pages 609--616, 2009.

\bibitem{TIMIT-89}
K.-L. Lee and H.-W. Hon.
\newblock Speaker-independent phone recognition using hidden {M}arkov models.
\newblock {\em IEEE Trans. Acoust., Speech, Signal Process.},
  37(11):1641--1648, 1989.

\bibitem{Pan2012}
J.~Pan, C.~Liu, Z.~Wang, Y.~Hu, and H.~Jiang.
\newblock Investigations of deep neural networks for large vocabulary
  continuous speech recognition: Why {DNN} surpasses {GMM}s in acoustic
  modelling.
\newblock In {\em Proc. of International Symposium on Chinese Spoken Language
  Processing}, 2012.

\bibitem{RBFNetwork1991}
J.~Park and I.~W. Sandberg.
\newblock Universal approximation using radial-basis-function networks.
\newblock {\em Neural Computation}, 3(2):246--257, 1991.

\bibitem{EdmundRolls98}
E.~T. Rolls and A.~Treves.
\newblock {\em Neural Networks and Brain Function}.
\newblock Oxford University Press, 1998.

\bibitem{Roweis98}
S.~Roweis.
\newblock {EM} algorithms for {PCA} and {SPCA}.
\newblock In {\em Advances in Neural Information Processing Systems (NIPS)},
  pages 626--632, 1998.

\bibitem{SainathLowRank13}
T.~Sainath, B.~Kingsbury, V.~Sindhwani, E.~Arisoy, and B.~Ramabhadran.
\newblock Low-rank matrix factorization for deep neural network training with
  high-dimensional output targets.
\newblock In {\em Proceedings of IEEE International Conference on Acoustic,
  Speech, Signal Processing (ICASSP 2013)}, 2013.

\bibitem{Tipping_1999}
M.~E. Tipping and C.~M. Bishop.
\newblock Mixtures of probabilistic principal component analyzers.
\newblock {\em Neural computation}, 11(2):443--482, 1999.

\bibitem{Tipping99b}
M.~E. Tipping and C.~M. Bishop.
\newblock Probabilistic principle component analysis.
\newblock {\em Journal of the Royal Statistical Society}, Series B
  21(3):611--622, 1999.

\bibitem{Vincent-2008}
P.~Vincent, H.~Larochelle, Y.~Bengio, and P.~Manzagol.
\newblock Extracting and composing robust features with denoising autoencoders.
\newblock In {\em Proceedings of the 25th international conference on Machine
  learning (ICML)}, pages 1096--1103, 2008.

\bibitem{JianXueMSR13}
J.~Xue, J.~Li, and Y.~Gong.
\newblock Restructuring of deep neural network acoustic models with singular
  value decomposition.
\newblock In {\em Proceedings of Interspeech 2013}, 2013.

\bibitem{XueShaofei2014}
S.~Xue, O.~Abdel-Hamid, H.~Jiang, L.~Dai, and Q.~Liu.
\newblock Fast adaptation of deep neural network based on discriminant codes
  for speech recognition.
\newblock {\em IEEE/ACM Trans. on Audio, Speech and Language Processing},
  22(12):1713--1725, 2014.

\end{thebibliography}
\bibliographystyle{plain}

\newpage 

\begin{center}
\LARGE \bf Appendix
\end{center}

\appendix

\section{Learning HOPE when $\hat{\bf U}$ is not orthonormal}

\label{appendix-non-orthonormal-U}

In some tasks, in additional to dimension reduction, we may want to use the projection matrix ${\bf U}$ to perform signal whitening to ensure the signal projection 
${\bf z}$ has roughly the same variance along all dimensions. This has been shown to be quite important for many image recognition and speech recognition tasks. In this case, we may still want to impose the orthogonal constraints among all row vectors of ${\bf U}$, i.e., ${\bf u}_i \cdot {\bf u}_j = 0 \;\; (i,j = 1, \cdots, M, \; i\neq j)$, but these row vectors may not be of unit length, i.e., $|{\bf u}_i| \geq 1 \;\; (i=1,\cdots,M)$. Moreover, it is better not to  whiten the residual noises in the remaining $D-M$ dimensions to amplify them unnecessarily. Therefore, we still enforce the unit-length constraints for the matrix ${\bf V}$, i.e., $|{\bf u}_i| = 1 \; (i=M+1,\cdots,D)$. Because ${\bf U}$ is not orthonormal anymore, when we compute the likelihood function of the original data in  eq.(\ref{eq-HOPE-likelihood-WithoutJacobian}) for HOPE, we have to include the Jacobian term as follows:

\begin{eqnarray}  \label{eq-HOPE-likelihood-WithJacobian}
& & {\cal L}({\bf U}, {\boldsymbol \Theta}, \sigma \; | \; {\bf X}) = 
 \sum_{n=1}^N \; \ln \Pr({\bf x}_n) = \sum_{n=1}^N  \bigg[ \ln |\hat{\bf U}^{-1}| + \ln \Pr({\bf z}_n) + \ln \Pr({\bf n}_n) \bigg] \nonumber \\ 
& = & 
\underbrace{ N\cdot  \ln |\hat{\bf U}^{-1}| }_{{\cal J}({\bf U})} + 
\underbrace{ \sum_{n=1}^N \; \ln \left(\sum_{k=1}^K  \;\; \pi_k \cdot f_k( {\bf U}  {\bf x}_n | {\boldsymbol \theta}_k) \right)}_{{\cal L}_1({\bf U},{\boldsymbol \Theta})}
+ \underbrace{\sum_{n=1}^N \;  \ln \bigg( {\cal N}\big( {\bf n}_n \, | \, {\bf 0}, \sigma^2 {\bf I}     \big)   \bigg)}_{{\cal L}_2({\bf U}, \sigma)} \nonumber \\
\end{eqnarray}

Because ${\bf U}{\bf U}^T \neq {\bf I}$ here, we have
\begin{equation}
{\bf x}_n = 	{\bf U}^T ({\bf U} {\bf U}^T)^{-1} {\bf z}_n + {\bf V}^T {\bf n}_n
\end{equation}
and then we have
\begin{eqnarray}
{\bf x}_n^T {\bf x}_n & = & \bigg({\bf n}_n^T {\bf V} + {\bf z}_n^T ({\bf U} {\bf U}^T)^{-1} {\bf U} \bigg)	
\bigg( {\bf V}^T {\bf n}_n + {\bf U}^T ({\bf U} {\bf U}^T)^{-1} {\bf z}_n \bigg) \nonumber \\
& = & {\bf n}_n^T {\bf n}_n + {\bf z}_n^T ({\bf U} {\bf U}^T)^{-1}  {\bf z}_n 
 =  {\bf n}_n^T {\bf n}_n + {\bf x}_n^T {\bf U}^T ({\bf U} {\bf U}^T)^{-1}  {\bf U} {\bf x}_n
\end{eqnarray}

Therefore, we may derive the residual noise energy as:
\begin{eqnarray}
{\bf n}_n^T {\bf n}_n & = & 
{\bf x}_n^T {\bf x}_n - {\bf x}_n^T {\bf U}^T ({\bf U} {\bf U}^T)^{-1}  {\bf U} {\bf x}_n
\nonumber \\
& = &  {\bf x}_n^T \bigg[ {\bf I} -  {\bf U}^T  ({\bf U} {\bf U}^T)^{-1} {\bf U}   \bigg] {\bf x}_n
\end{eqnarray}

In this case, ${\cal L}_2({\bf U}, \sigma)$ can be expressed as:
\begin{equation}
{\cal L}_2({\bf U}, \sigma) = - \frac{N}{2} \ln (\sigma^2) - \frac{1}{2 \sigma^2} \sum_{n=1}^N \;
\bigg[ {\bf x}_n^T {\bf x}_n - {\bf x}_n^T  {\bf U}^T  ({\bf U} {\bf U}^T)^{-1} {\bf U}  {\bf x}_n \bigg]
\end{equation}

Therefore, its gradient with respect to ${\bf U}$, can be derived as follows:

\begin{eqnarray}
\frac{\partial {\cal L}_2({\bf U}, \sigma)}{\partial {\bf U}} & = &  \frac{1}{\sigma^2}  \sum_{n=1}^N \;
\bigg[ ({\bf U} {\bf U}^T)^{-1} {\bf U} {\bf x}_n  {\bf x}_n^T   
- ({\bf U} {\bf U}^T)^{-1} {\bf U} {\bf x}_n  {\bf x}_n^T {\bf U}^T ({\bf U} {\bf U}^T)^{-1} {\bf U}  \bigg] \nonumber \\
& = &  \frac{1}{\sigma^2}  \sum_{n=1}^N \;\;
({\bf U} {\bf U}^T)^{-1} {\bf U} {\bf x}_n  {\bf x}_n^T  \bigg[ {\bf I}  - {\bf U}^T ({\bf U} {\bf U}^T)^{-1} {\bf U} \bigg]
\end{eqnarray}

Next, we consider the Jacobian term, ${\cal J}({\bf U})$, which can be computed as follows:
\begin{equation}
{\cal J}({\bf U}) = N\cdot  \ln |\hat{\bf U}^{-1}| 
= - N \sum_{i=1}^M  \; \ln |{\bf u}_i|
\end{equation}

Because of 
$
\frac{\partial {\cal J}({\bf U})}{\partial {\bf u}_i} = - \frac{N\cdot {\bf u}_i}{ |{\bf u}_i|^2} \;
(i=1, \cdots, M)$, it is easy to show that its derivative with respect to ${\bf U}$ can be derived as:
\begin{equation}
\frac{\partial {\cal J}({\bf U})}{\partial {\bf U}} =
-N \cdot \big( {\bf U} {\bf U}^T\big)^{-1} {\bf U}. 
\end{equation}

Similarly, the HOPE parameters, i.e., ${\bf U}$, ${\boldsymbol \Theta}$ and $\sigma$, can be estimated by maximizing the above likelihood function as follows:
\begin{equation}
\{ {\bf U}^*, {\boldsymbol \Theta}^*, \sigma^*\}	  = {\arg\max}_{ {\bf U}, {\boldsymbol \Theta}, \sigma } \;\;\; 
{\cal L}({\bf U}, {\boldsymbol \Theta}, \sigma \; | \; {\bf X})
\end{equation}
subject to an orthogonal constraint: 
\begin{equation} \label{eq-HOPE-orthogonal-constraint}
{\bf U} {\bf U}^{T}  = {\boldsymbol \Phi},
\end{equation}
where ${\boldsymbol \Phi}$ is a diagonal matrix. The above constraint can also be 
implemented as an penalty term similar to eq.(\ref{eq-HOPE-orthogonal-constraint-term}).
However, the norm of each row vector is relaxed as follows:
\begin{equation}
|{\bf u}_i| \geq 1   \;\;\; (i=1,\cdots, M)
\end{equation}

The log-likelihood function related to the signal model, ${\cal L}_1({\bf U}, {\boldsymbol \Theta})$, and signal variance, $\sigma^2$, are calculated in the same way as before.

\section{Derivatives of movMFs}
\label{appendix_derivatives_movMF}

The partial derivatives of the objective function in eq.(\ref{moVMF_HOPE}) w.r.t  all $\boldsymbol{\mu}_k$ can be computed as follows:
\begin{equation}\label{App_1}
     \frac{{\partial {\cal L}_1 ({\bf U}, {\boldsymbol \Theta})}}{{\partial {\boldsymbol \mu _k}}} = \sum_{n=1}^{N} \frac{\pi_k\cdot \bigg({\cal C}_M^{'}(|\boldsymbol{\mu}_k|)\frac{\partial \kappa_k}{\boldsymbol{\mu}_k}\cdot e^{{\bf z}_n^{T}\cdot \boldsymbol{\mu}_k}+{\cal C}_M(|\boldsymbol{\mu}_k|)\cdot e^{{\bf z}_n^{T}\cdot \boldsymbol{\mu}_k}\cdot{\bf z}_n \bigg)}{\sum_{j=1}^{K}\pi_j\cdot{\cal C}_{M} (|\boldsymbol{\mu}_j|) \cdot e^{{\bf z} \cdot {\boldsymbol \mu}_j}}
\end{equation}
where we have  
\begin{equation}\label{App_2}
\frac{\partial \; |\boldsymbol{\mu}_k |}{\partial \boldsymbol{\mu}_k}=\frac{\boldsymbol{\mu}_k}{|\boldsymbol{\mu}_k|}.
\end{equation}
As to ${\cal C}_M^{'}(\kappa)$, for brevity, let us denote $s=\frac{M}{2}-1$, and $\xi = (2\pi)^{s+1}$.
\begin{equation}
\begin{gathered}
  {{\cal C}_M^{'}}(\kappa ) = \frac{{s \cdot {\kappa ^{s - 1}}}}{{\xi {I_s}(\kappa )}} - \frac{{{\kappa ^s} \cdot {{I'}_s}(\kappa )}}{{\xi {I_s}^2(\kappa )}} = \frac{{{\kappa ^s}}}{{\xi {I_s}(\kappa )}} \bigg(\frac{s}{\kappa } - \frac{{{{I'}_s} (\kappa )}}{{{I_s}(\kappa )}} \bigg) \hfill \\
  \qquad \quad  = {{\cal C}_M}(\kappa ) \cdot \bigg(\frac{s}{\kappa } - \frac{{{{I'}_s}(\kappa )}}{{{I_s}(\kappa )}} \bigg) \hfill \\ 
\end{gathered} 
\end{equation}
where $I_v(\cdot)$ denotes the Bessel function of the first kind at the order $v$. Because we have
\[\kappa {I_{s + 1}}(\kappa ) = \kappa {I'_s}(\kappa ) - s{I_s}(\kappa ) \Rightarrow \frac{s}{\kappa } - \frac{{{{I'}_s}(\kappa )}}{{{I_s}(\kappa )}} =  - \frac{{{I_{s + 1}}(\kappa )}}{{{I_s}(\kappa )}}\]
Thus, we may derive  
\begin{equation}\label{App_3}
 {{\cal C}'_M}(\kappa ) =  - {{\cal C}_M}(\kappa ) \cdot \frac{{{I_{s + 1}}(\kappa )}}{{{I_s}(\kappa )}}
\end{equation}

Substituting  eq. (\ref{App_2}) and eq. (\ref{App_3}) into  eq. (\ref{App_1}),  we obtain the partial derivative of the objective function in eq. (\ref{moVMF_HOPE}) w.r.t  $\boldsymbol{\mu}_k$ as follows:
\begin{equation} \label{eq-vMF-derivatives-mu}
\begin{gathered}
\frac{{\partial {\cal L}_1 ({\bf U}, {\boldsymbol \Theta})}}{{\partial {\boldsymbol{\mu}_k}}} = \mathop \sum \limits_{n = 1}^N \frac{{{\pi _k} \cdot {{\cal C}_M}({|\boldsymbol{\mu}_k |}) \cdot {e^{{{\mathbf{z}}_n} \cdot {\boldsymbol{\mu}_k}}} \bigg({{\mathbf{z}}_n} - \frac{{{\boldsymbol{\mu} _k}}}{{{|\boldsymbol{\mu}_k |}}} \cdot \frac{{{I_{M/2}}({|\boldsymbol{\mu}_k |} )}}{{{I_{M/2 - 1}}({|\boldsymbol{\mu}_k |})}} \bigg)}}{{\sum\nolimits_{j = 1}^K \pi_j\cdot{{\cal{C}_M}} ({|\boldsymbol{\mu}_j |}) \cdot {e^{{{\mathbf{z}}_n} \cdot {\boldsymbol{\mu} _j}}}}} \hfill \\
\qquad  = \sum\limits_{n = 1}^N \gamma  ({z_{nk}}) \cdot \bigg( {{\mathbf{z}}_n} - \frac{{{\boldsymbol{\mu}_k}}}{{{|\boldsymbol{\mu}_k |}}} \cdot \frac{{{I_{M/2}}({|\boldsymbol{\mu}_k |})}}{{{I_{M/2 - 1}}({|\boldsymbol{\mu}_k |})}} \bigg)  \hfill \\ 
\end{gathered} 
\end{equation}
where  
$ \gamma(z_{nk})=\frac{\pi_k\cdot {\cal C}_{M} (|\boldsymbol{\mu}_k |) \cdot e^{{\bf z}_n \cdot {\boldsymbol \mu}_k}}{\sum_{j=1}^{K}\pi_j\cdot{\cal C}_{M} (|\boldsymbol{\mu}_j |) \cdot e^{{\bf z}_n \cdot {\boldsymbol \mu}_j}}
$ is the occupancy statistics of $k$-th component of ${\bf z}_n$.

Next, let us consider the
partial derivative of the objective function in eq.(\ref{moVMF_HOPE}) w.r.t  ${\bf U}$.
Based on the chain rule, we have 
\begin{equation}\label{eq.W_1}
\frac{{\partial {\cal L}_1 ({\bf U}, {\boldsymbol \Theta})}}{{\partial {\mathbf{U}}}} = \frac{{\partial {\cal L}_1 ({\bf U}, {\boldsymbol \Theta})}}{{\partial \widetilde {\mathbf{z}}_n}} \cdot \frac{{\partial \widetilde {\mathbf{z}}_n}}{{\partial {\mathbf{U}}}} = \bigg(\frac{{\partial {{\mathbf{z}}_n^T}}}{{\partial {\mathbf{\tilde z}}_n}} \cdot \frac{{\partial {\cal L}_1 ({\bf U}, {\boldsymbol \Theta})}}{{\partial {\mathbf{z}}_n}} \bigg) \cdot \frac{{\partial \widetilde {\mathbf{z}}_n}}{{\partial {\mathbf{U}}}}
\end{equation}

Furthermore, we may derive
\begin{equation}\label{eq.W_2}
 \frac{\partial {\cal L}_1 ({\bf U}, {\boldsymbol \Theta})}{\partial {\bf z}_n} = \sum_{n=1}^{N} \sum_{k=1}^{K}\frac{\pi_k\cdot {\cal C}_{M} (|\boldsymbol{\mu}_k |) \cdot e^{{\bf z}_n \cdot {\boldsymbol \mu}_k}}{\sum_{j=1}^{K}\pi_j\cdot{\cal C}_{M} (|\boldsymbol{\mu}_j |) \cdot e^{{\bf z}_n \cdot {\boldsymbol \mu}_j}}\cdot \boldsymbol{\mu}_k =\sum_{n=1}^{N} \sum_{k=1}^{K} \gamma(z_{nk})\cdot \boldsymbol{\mu}_k
\end{equation}
\begin{equation}
\begin{gathered}\label{eq.W_3} 
\frac{\partial {\bf z}_n^T}{\partial {\tilde {\bf z}_n}}  = \frac{\partial { (\tilde{\bf z}_n^T/|\tilde{\bf z}_n|)}}{\partial {\tilde {\bf z}_n}} = \frac{1}{|\tilde{\bf z}_n|^2}\bigg(\frac{\partial \tilde{\bf z}_n^T}{\partial \tilde{\bf z}_n} |\tilde{\bf z}_n|-\frac{\partial |\tilde{\bf z}_n|}{\partial \tilde{\bf z}_n} \tilde{\bf z}_n^T \bigg)\\
=\frac{1}{|\tilde{\bf z}_n|}\bigg({\bf I} - \frac{\tilde{\bf z}_n\tilde{\bf z}_n^T}{|\tilde{\bf z}_n|^2}\bigg)=\frac{1}{|\tilde{\bf z}_n|} \bigg( {\bf I} - {\bf z}_n{\bf z}_n^T \bigg)\\
\end{gathered}
\end{equation}

Substituting  eq.(\ref{eq.W_2}) and eq. (\ref{eq.W_3}) into  eq. (\ref{eq.W_1}),  we can obtain
\begin{equation}
\frac{{\partial {\cal L}_1 ({\bf U}, {\boldsymbol \Theta})}}{{\partial {\mathbf{U}}}}  = \sum_{n=1}^{N} \sum_{k=1}^{K} \frac{\gamma(z_{nk})}{|\tilde{\bf z}_n|}({\bf I} - {\bf z}_n{\bf z}_n^T) \boldsymbol{\mu}_k {\bf x}_n^{T}
\end{equation}

\section{Numerical Methods for $I_v(\cdot)$}
\label{appendix-numerical-Bessel}

In the learning algorithm for movMFs, we may need to compute the Bessel functions, $I_v(\cdot)$, in several places. 
First of all, we need to compute the normalization term ${\cal C}_M(\kappa)$ when calculating the likelihood function of a vMF distribution as in eq.(\ref{eq-movMF-pdf}). Secondly, we need to calculate the rations of the modified Bessel functions, $A_d(\kappa)=\frac{I_{M/2}(\kappa)}{I_{M/2-1}(\kappa)}$, in eq.(\ref{eq-vMF-derivatives-mu}).
As we know, the modified Bessel functions of the first kind take the following form:
 \begin{equation}\label{bessel}
 I_{d}(\kappa) = \sum_{k\geq 0} \frac{1}{\Gamma(d+1+k)k!}(\frac{\kappa}{2})^{2k+d},
 \end{equation}

From eq. (\ref{bessel}), we can see that when $\kappa \gg d$, $I_{d}(\kappa) $ overflows quite rapidly. Meanwhile,  
when $\kappa = o(d) $ and $d \rightarrow \infty$, $I_{d}(\kappa) $ underflows quite rapidly. In this work,
 we use the approximation strategy in eq. (9.7.7) on page 378 of \cite{Abramowitz_1964} as follows:

\begin{equation}\label{Approx_bessel}
 I_{d}(\kappa) \sim \frac{1}{\sqrt{2\pi d}}\cdot \frac{e^{d \eta }}{(1+(\frac{\kappa}{d})^2)^{1/4}}\cdot \bigg[ 1+\sum_{k=1}^{\infty}\frac{u_k(t)}{d^{k}} \bigg]
 \end{equation}
where we have 
\begin{equation}  \label{eq-approximation-t}
t=\frac{1}{\sqrt{1+(\kappa/d)^2}}
\end{equation}
\begin{equation}  \label{eq-approximation-eta}
\eta = \sqrt{1+(\kappa/d)^2} + \ln \frac{\kappa/d}{1+\sqrt{1+(\kappa/d)^2}} 
\end{equation}
with the functions $u_k(t)$ taking the following forms:
 \[\begin{gathered}
  {u_0}(t) = 1 \hfill \\
  {u_1}(t) = (3t - 5{t^3})/24 \hfill \\
  {u_2}(t) = (81{t^2} - 462{t^4} + 385{t^6})/1152 \hfill \\ 
\end{gathered} \]

Refer to page 366 of \cite{Abramowitz_1964} for other  higher orders $u_k(t)$. 

Usually, the sum of the term $[1+\sum_{k=1}^{\infty}\frac{u_k(t)}{d^{k}}]$ in eq.(\ref{Approx_bessel}) is very small and it is safe to eliminate it from evaluation in most cases. Then, after substituting eq.(\ref{eq-approximation-t}) and eq.(\ref{eq-approximation-eta}) into eq.(\ref{Approx_bessel}), the logarithm of the approximated modified Bessel function is finally computed  as follows:
\begin{equation}  \label{eq-Bessel-approximation}
\ln I_d(\kappa)= -\ln \sqrt{2\pi d } + d\cdot \bigg(\sqrt{1+(\kappa/d)^2} +\ln \frac{\kappa}{d} -\ln \Big( 1+\sqrt{1+(\kappa/d)^2} \Big)  \bigg) -\frac{1}{4}\ln \Big(1+(\frac{\kappa}{d})^2 \Big)
\end{equation}

In this work, the approximation in eq.(\ref{eq-Bessel-approximation}) is used to compute all Bessel functions in the learning algorithms for movMFs. 

\end{document}